%% file: main.tex
\documentclass{article}
\input{math_commands}

\usepackage{amsmath,amsfonts,bm}
\usepackage{caption}
\usepackage{subcaption}
\usepackage{verbatim}
\usepackage[dvipsnames]{xcolor}
\definecolor{orange}{rgb}{0.93725,0.52549,0.2117647}
\definecolor{blue}{rgb}{0.23137,0.4588,0.68627}
\usepackage{wrapfig}
\usepackage{dirtytalk}
\usepackage{wrapfig}
\usepackage{color, colortbl}
\usepackage{microtype}
\usepackage{stfloats}
\usepackage{graphicx}
\usepackage{booktabs} % for professional tables
\usepackage{multirow} % tables with cell spanning multiple rows
\usepackage{siunitx}
\usepackage{array}
\usepackage{xcolor} 
\newcommand{\thickhline}{%
    \noalign {\ifnum 0=`}\fi \hrule height 1pt
    \futurelet \reserved@a \@xhline
}
\newcolumntype{"}{@{\hskip\tabcolsep\vrule width 1pt\hskip\tabcolsep}}

\usepackage{upquote,textcomp,amsmath}

\usepackage[pagebackref=false,breaklinks=true,letterpaper=true,colorlinks,bookmarks=false]{hyperref}
\hypersetup{colorlinks,breaklinks,
            urlcolor=[rgb]{0.93725,0.52549,0.2117647},
            citecolor=[rgb]{0.23137,0.4588,0.68627},
            linkcolor=[rgb]{0.93725,0.52549,0.2117647}}

\usepackage{amsthm}
\renewcommand{\eqref}[1]{Eq.~\ref{#1}}

\usepackage{xcolor}
\usepackage{authblk}

\usepackage{iclr2027_conference,times}
\iclrfinalcopy

\usepackage[utf8]{inputenc} % allow utf-8 input
\usepackage[T1]{fontenc}    % use 8-bit T1 fonts
\usepackage{hyperref}       % hyperlinks
\usepackage{url}            % simple URL typesetting
\usepackage{booktabs}       % professional-quality tables
\usepackage{amsfonts}       % blackboard math symbols
\usepackage{nicefrac}       % compact symbols for 1/2, etc.
\usepackage{microtype}      % microtypography
\usepackage{longtable}

\title{Multitask Interactions in Recurrent Dynamics across 100 Cognitive Tasks}

\author{
  \textbf{Sílvia García-Hernández}$^{1,2,3}$, \textbf{Leema Hamid}$^{1,2,3}$, \textbf{Manuel Brenner}$^{1,2,3}$ \\
  $^1$Ernst Strüngmann Institute (ESI) of the Max Planck Society, Germany\\ 
  $^2$Goethe University (GU) Frankfurt, University Hospital, Department for Psychiatry, Psychosomatic\\
  Medicine and Psychotherapy, Germany \\
  $^3$GU Frankfurt, Cooperative Brain Imaging Center, Germany \\
  Corresponding authors: \texttt{\{silvia.garcia,  manuel.brenner\}@esi-frankfurt.de} \\
}

\begin{document}

\maketitle
\lhead{Preprint. Under review.}

\begin{abstract}

The brain continually learns tasks with very different computational demands. Recurrent neural networks (RNNs) have been used as dynamical surrogates for studying task solutions, with recent work extending this approach to joint training on multitask batteries. However, it remains unclear how tasks interact as these batteries grow in scale and heterogeneity: under what conditions do they transfer or interfere with one another, and how do these effects relate to task complexity and model structure?
We study this question by simultaneously training hierarchical Almost-Linear RNNs (AL-RNNs) on 100 tasks spanning a wide range of cognitive demands. AL-RNNs allow us to vary the amount of nonlinearity in the recurrence, while hierarchical task features provide a low-dimensional interface for task-specific adaptation within shared dynamics.
On this task battery, performance improves with both nonlinear capacity and task-feature dimensionality, with the joint hierarchical model outperforming both individually trained models and a single unstructured joint model. Many tasks are individually solvable with linear dynamics, yet cannot be solved together by a shared linear model.
Through hierarchical task features, task-specific linear region usage, and task geometry, the trained model provides several related but distinct measures of task similarity that capture how computations are separated and combined within the shared backbone. Across a series of experiments, we relate these representations to individual task complexity, joint solvability, pairwise transfer learning, and leave-one-out adaptation. Together, our experiments characterize task interactions in multitask recurrent learning while demonstrating both the potential and the challenges of reusable dynamical backbones for cognitive computation.

\end{abstract}

\section{Introduction}

The mammalian brain is a remarkably flexible learning system, acquiring a large repertoire of cognitive tasks by both efficiently leveraging shared structure across tasks (transfer learning) while preventing new learning from overwriting existing solutions (catastrophic forgetting).
Achieving this requires a dynamical substrate supporting reusable motifs and task-specific separation. However, isolating and interpreting these mechanisms from recordings of complex nonlinear brain dynamics remains challenging. To better understand the dynamical mechanism underpinning task solutions, task-trained recurrent neural networks (RNNs) have emerged as dynamical surrogate models that allow researchers to analyze the resulting attractor dynamics and compare them to neural recordings \citep{mante_context-dependent_2013, sussillo_opening_2013}. Historically focused on single-task settings, this approach has identified task-specific motifs but leaves open how they coexist when multiple tasks are embedded in a shared system.

In domains like vision and language, where data modalities are structurally homogeneous, positive transfer is often the default expectation \citep{raffel_exploring_2020}. However, cognitive tasks vary substantially in input structure, delay periods, and computational demands, creating a real risk of interference, requiring the model to actively separate conflicting computations within the shared dynamics. Recent work has begun to compile task batteries and investigated joint training and dynamical mechanisms \citep{yang_task_2019,turner_simplicity_2023, driscoll_flexible_2024}. So far, these task suites have usually been relatively small and structured, with many of their tasks admitting surprisingly simple solutions \citep{khona2023winninglotteryneuralconnectivity}. This motivates larger and more heterogeneous batteries \citep{khona2023winninglotteryneuralconnectivity} for asking how task interactions scale when the shared model must accommodate diverse computational demands.

Understanding these interactions requires a framework that can explicitly control the capacity for nonlinear computation and task-specific adaptation. In this work, we employ hierarchical Almost-Linear RNNs (AL-RNNs) as an interpretable experimental framework to study multitask interactions within a shared recurrent dynamical backbone. The AL-RNN architecture allows us to vary the amount of piecewise-linear nonlinear computation, while hierarchical task features provide a low-dimensional interface for task-specific adaptation. We train this model jointly on a heterogeneous battery of 100 supervised cognitive tasks, combining paradigms adapted from prior multitask studies \citep{yang_task_2019,driscoll_flexible_2024,khona2023winninglotteryneuralconnectivity} with additional tasks compiled from the cognitive and neuroscientific literature. We compare joint hierarchical models with individually trained networks and flat joint models to characterize how heterogeneous tasks interact under shared recurrent dynamics. The hierarchical AL-RNN provides a parameter-efficient compromise that outperforms matched-size individual models and unstructured joint models. Across several analyses, we find that nonlinearity, low-dimensional task features, and recurrent state geometry provide complementary forms of task organization, while jointly learning more heterogeneous tasks requires increasingly complex models. These representations define interpretable task-similarity spaces that relate to transfer learning: some tasks systematically produce negative transfer when used as initializations, and a task's isolation in the learned space was predictive of poorer transfer and poorer leave-one-out adaptation to it. Together, these findings demonstrate that effective multitask computation is possible but requires actively structured capacity to manage interference within shared dynamics.

\section{Related Work}
Recurrent neural networks have become standard tools for modeling cognitive dynamics, successfully replicating neural activity and behavior in single-task settings \citep{mante_context-dependent_2013, sussillo_opening_2013, yang_task_2019}. While recent work has scaled cognitive modeling, e.g. using large language models fine-tuned on broad behavioral datasets \citep{binz_foundation_2025}, RNNs remain unique in providing interpretable, low-dimensional dynamical systems directly comparable to neural recordings \citep{khona_attractor_2022,durstewitz_reconstructing_2023}. However, most such models focus on isolated tasks, and how heterogeneous tasks interact within a single network remains poorly understood.

Recent work has increasingly examined recurrent networks trained on multiple cognitive tasks simultaneously. Task suites such as NeuroGym \citep{crocioni_neurogym_2026} and Mod-Cog \citep{khona2023winninglotteryneuralconnectivity} make such studies possible, while highlighting that many of their tasks admit simple solutions, from sparse or even diagonal connectivity to linear dynamics \citep{khona2023winninglotteryneuralconnectivity}. \citet{yang_task_2019} trained one RNN on 20 cognitive tasks and found clustered units and compositional task representations, and \citet{driscoll_flexible_2024} showed that related tasks reuse shared dynamical motifs. \citet{duncker_organizing_2020} organize dynamics into task-specific, orthogonal subspaces while reusing motifs across similar computations. \citet{turner_simplicity_2023} call this tendency of RNNs to develop shared common representations across tasks a "simplicity bias", linking it to the emergence of shared attractor structure during training. \citet{flesch_orthogonal_2022} link such orthogonal, low-dimensional representations to robustness across contexts in networks and brains. Together, this literature has mainly characterized how structured task relationships enable sharing and compositionality. 

Indeed, where tasks are structurally homogeneous, sharing parameters across tasks typically helps \citep{raffel_exploring_2020}. However, sharing is not necessarily beneficial, and in multitask settings with heterogeneous tasks, these can compete for shared capacity and can degrade one another, leading to the concept of “negative transfer”, which is well established across the wider multitask learning literature \citep{standley_which_2020,zhang_survey_2023}, but has received less systematic attention in dynamical models of cognitive task batteries. One line of work treats such interference as an optimization problem of conflicting gradient directions between tasks \citep{yu_gradient_2020, sener_multi-task_2018}, though the need for such specialized optimizers is contested \citep{kurin_defense_2022}. A second line of work addresses interference through model structure, explicitly balancing shared and task-specific computation, with modularity constituting a useful inductive bias \citep{goyal_recurrent_2020}.
A number of approaches separate shared from task-specific computation explicitly, e.g. through learned routing between function blocks \citep{rosenbaum_routing_2018} or task-specific gates over shared experts \citep{ma_modeling_2018}, while a number of other approaches have observed these structures emerge in trained networks \citep{yang_task_2019}, and leveraged in neural populations \citep{dubreuil_role_2022}.

Task conditioning provides another mechanism for combining sharing and separation, and can create a shared representational geometry that supports generalization to unseen tasks \citep{riveland_natural_2024}. Task identity can be characterized during joint training \citep{fifty_efficiently_2021} or post hoc, for instance through Fisher-information-based embeddings \citep{achille_task2vec_2019}. Hypernetworks generate model parameters from task features and can mitigate forgetting in continual learning \citep{oswald_continual_2019,ehret_continual_2020,durstewitz_what_2025}, while hierarchical models combine low-dimensional task-specific features with shared parameters \citep{brenner_learning_2024}. These learned representations also define task spaces in which similarity can be quantified and related to transfer, providing a bridge between structured multitask models and the study of task interactions.

\section{Methods}

Studying multitask interactions requires a model that allows us to quantify task complexity, distinguish shared from task-specific computation, and remain tractable as the number and heterogeneity of tasks increase. It should also provide an adaptation interface that allows new tasks to be learned without fitting an independent network for each task or modifying the shared dynamics. These requirements motivate a model that is efficient in both latent dimensionality and dynamical complexity, while retaining sufficient task-specific flexibility to avoid interference. We combine three components to meet these requirements: i) an Almost-Linear RNN (AL-RNN) \citep{brenner_almost-linear_2024}, in which latent dimension $M$ and nonlinearity $P$ are decoupled and can be tuned independently (Sec.~\ref{sec:alrnn}), ii) a hierarchical inference framework \citep{brenner_learning_2024}, which generates task-specific recurrent parameters from low-dimensional task features and shared group-level parameters, providing a common dynamical backbone with an interpretable task interface (Sec.~\ref{sec:hierarchical}), and iii), a shared input-output interface that accommodates heterogeneous task formats while keeping the recurrent parameterization consistent across all tasks (Appx. \ref{app:task_suite}).

\subsection{Almost-linear RNNs} \label{sec:alrnn}
The AL-RNN allows the two principal capacity axes (latent dimensionality $M$ and the number of nonlinear units $P$) to be varied independently. This allows us to distinguish tasks that require rich state-space structure despite being linearly solvable from tasks that require nonlinear computation even in relatively simple geometries \citep{brenner_uncovering_2026}. The AL-RNN evolves a latent state $\bm{z}_t \in \mathbb{R}^{M}$ according to
\begin{align}\label{eq:alrnn}
    \bm{z}_t = \bm{A}\bm{z}_{t-1} + \bm{W}\,\Phi^{*}(\bm{z}_{t-1}) + \bm{C}\bm{s}_t + \bm{h},
\end{align}
with diagonal $\bm{A}\in\mathbb{R}^{M\times M}$, off-diagonal $\bm{W}\in\mathbb{R}^{M\times M}$, input matrix $\bm{C}\in\mathbb{R}^{M\times N}$, bias $\bm{h}\in\mathbb{R}^{M}$, and external input $\bm{s}_t$. Nonlinearity is applied to the last $P \leq M$ units only,
\begin{align}\label{eq:phistar}
    \Phi^{*}(\bm{z}_t) = \left[z_{1,t},\,\ldots,\,z_{M-P,t},\,
    \phi(z_{M-P+1,t}),\,\ldots,\,\phi(z_{M,t})\right]^{\!\top},
\end{align}
with $\phi(z) = \max(0, z)$ as in the original formulation. For $P = 0$ the model is a linear state space model, while for $P = M$ it reduces to a standard piecewise-linear RNN \citep{durstewitz_state_2017}. The latent states are read out using a linear layer $\bm{B}\in\mathbb{R}^{5\times M}$ with $\bm{y}=\bm{B}\bm{z}$, with task-specific masks selecting the relevant timesteps and readout channels.

The piecewise-linear structure makes the model interpretable in terms of its linear region usage. For a given activation pattern $\bm{b}_t\in\{0,1\}^{P}$, where $b_{j,t} = \mathbf{1}\!\left\{z_{M-P+j,t}>0\right\}$, the nonlinear units contribute the diagonal gating matrix $\widetilde{\bm{D}}(\bm{b}_t)=\operatorname{diag}(\bm{b}_t)$, where $\widetilde{\bm{D}}(\bm{b}_t)\in\mathbb{R}^{M\times M}$ is diagonal, with zeros in the first $M-P$ entries and $\bm{b}_t$ in the final $P$ entries. Within a fixed region, the dynamics therefore reduce to the linear state-space system
$\bm{z}_t =\left(\bm{A}+\bm{W}\widetilde{\bm{D}}(\bm{b}_t)\right)\bm{z}_{t-1}+\bm{C}\bm{s}_t+\bm{h}$. Each binary activation pattern, i.e. its bitcode, such as $\bm{b}_t=(0,1,1,\ldots,0)$, defines one of the $2^P$ piecewise-linear regions. The bitcodes visited during a task then encode how the model uses nonlinearity to solve the task \citep{brenner_uncovering_2026}.

\subsection{Hierarchical Multitask Inference} \label{sec:hierarchical}

\begin{figure}[!htb]
    \centering
    \includegraphics[width=0.99\linewidth]{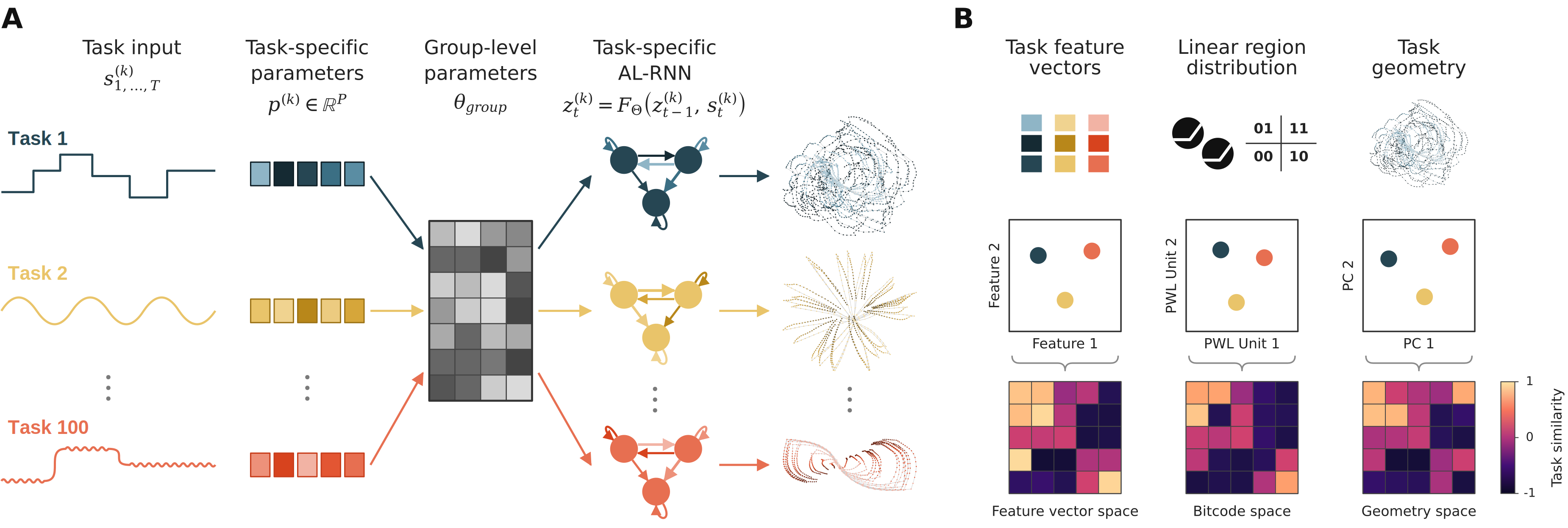}
    \caption{Hierarchical task conditioning and task-similarity spaces. \textbf{A}: Every task receives low-dimensional task features that modulate group-level parameters, producing task-conditioned AL-RNNs within a shared recurrent backbone. \textbf{B}: The learned model supports complementary task representations based on task features, piecewise-linear region usage, and state geometry.}
    \label{fig:framework}
\end{figure}

A joint model should share structure wherever tasks use related computations, while retaining enough task-specific flexibility to prevent interference. To achieve this, we use a hierarchical formulation \citep{brenner_learning_2024} that places task-specific variation in a low-dimensional space while generating the parameters of a task-specific dynamical system (DS) model from shared group-level parameters, depicted in Fig. \ref{fig:framework}. Each task $k\in\{1,\ldots,K\}$ is assigned an embedding $\bm{p}^{(k)}\in\mathbb{R}^{N_{\mathrm{feat}}}$, with $N_{\mathrm{feat}}\ll\dim(\bm{\theta}_{\mathrm{DSR}})$, and a shared hypernetwork $G_{\bm{\theta}_{\mathrm{group}}}$ maps this embedding to the parameters of its task-specific DS, parameterized e.g. by an AL-RNN:
\begin{equation}\label{eq:hierarchical}
    \bm{\theta}_{\text{DSR}}^{(k)} = G_{\bm{\theta}_{\text{group}}}\!\left(\bm{p}^{(k)}\right),
    \qquad
    \bm{\theta}_{\text{DSR}}^{(k)} = \{\bm{A}^{(k)}, \bm{W}^{(k)}, \bm{C}^{(k)}, \bm{B}^{(k)},\bm{h}^{(k)}\}.
\end{equation}
 We use a linear mapping for $G$, so that all task-specific models lie in an $N_{\text{feat}}$-dimensional linear subspace of parameter space. Since the group parameters $\bm{\theta}_{\text{group}}$ receive gradients from every task, any dynamical structure that is useful more than once should be preferentially represented there, while the only degrees of freedom individual to a task are the $N_{\text{feat}}$ coordinates of $\bm{p}^{(k)}$. Task-specific adaptation is therefore confined to a low-dimensional subspace. This formulation also automatically supplies task identity, such that all task-specific structure the model discovers is localized in one object.

\subsection{Task suite}\label{tasks_suite}
We train on a heterogeneous battery of $K=100$ supervised cognitive tasks spanning a diverse set of requirements, such as working memory, decision-making, perception, attention, timing, reasoning, and motor response. The battery combines task paradigms and variants adapted from prior multitask cognitive RNN studies \citep{yang_task_2019, khona2023winninglotteryneuralconnectivity, driscoll_flexible_2024}, with additional tasks compiled from the cognitive-science and systems-neuroscience literature. Following the compositional construction of \citet{khona2023winninglotteryneuralconnectivity}, selected base tasks are expanded with modifiers that add delay, sequence generation, interval integration, or combinations thereof. Each trial may contain context, stimulus, delay, and response phases. The final battery contains $66$ base task identities and $34$ compositional variants, each treated as a separate task with its own task embedding. The complete task list, readout types, source references, and modifier definitions are provided in Appx.~\ref{tasks_list}.

All tasks are jointly trained using a common masked mean-squared-error objective (Eq. \ref{eq:loss}, see Appx. \ref{app:training_procedure} for details). Accuracy is evaluated separately using task-appropriate trial-level criteria: final-step accuracy for single-response tasks, response-step accuracy for reaction-time tasks, and the fraction of correctly scored timesteps for sequential tasks. Directional outputs are evaluated by angular error, while scalar and vector outputs use calibrated tolerances. The choice of tolerances and robustness analyses are described in Appx. ~\ref{app:task_suite}. We report both mean accuracy across tasks and the number of solved tasks, where a task is considered solved above an $80\%$ accuracy threshold.

\section{Results}

\subsection{Performance across the task battery}
We trained three model variants on the full battery of 100 cognitive tasks: i) the task-conditioned hierarchical AL-RNNs, ii) flat AL-RNNs (a single model trained on all tasks at once), and iii) individually trained AL-RNNs (one model per task), each with matched parameter budgets. We further compared to hierarchical variants of vanilla RNNs,  GRUs and LSTMs. These results are summarized in Table \ref{tab:main_performance} for $200$ samples per task. We find that hierarchical formulations of the AL-RNN, GRU and LSTM perform best, reaching around $96\%$ test accuracy, clearly outperforming both individually trained models, reaching $91\%$, and the flat model, performing worst at around $81\%$ accuracy. Thus, joint training provides a substantial sharing benefit, even though many tasks are already solvable in isolation, while the structure of the sharing is important in order to avoid the negative interference observed in the flat model. These effects were also largely conserved across different sample sizes, with overall worse performance for smaller sample sizes (Fig. \ref{fig:samples_vs_nonlinearity}).

\input{tables/table_main_slim}

\begin{figure}[!htb]
    \centering
    \includegraphics[width=0.99\linewidth]{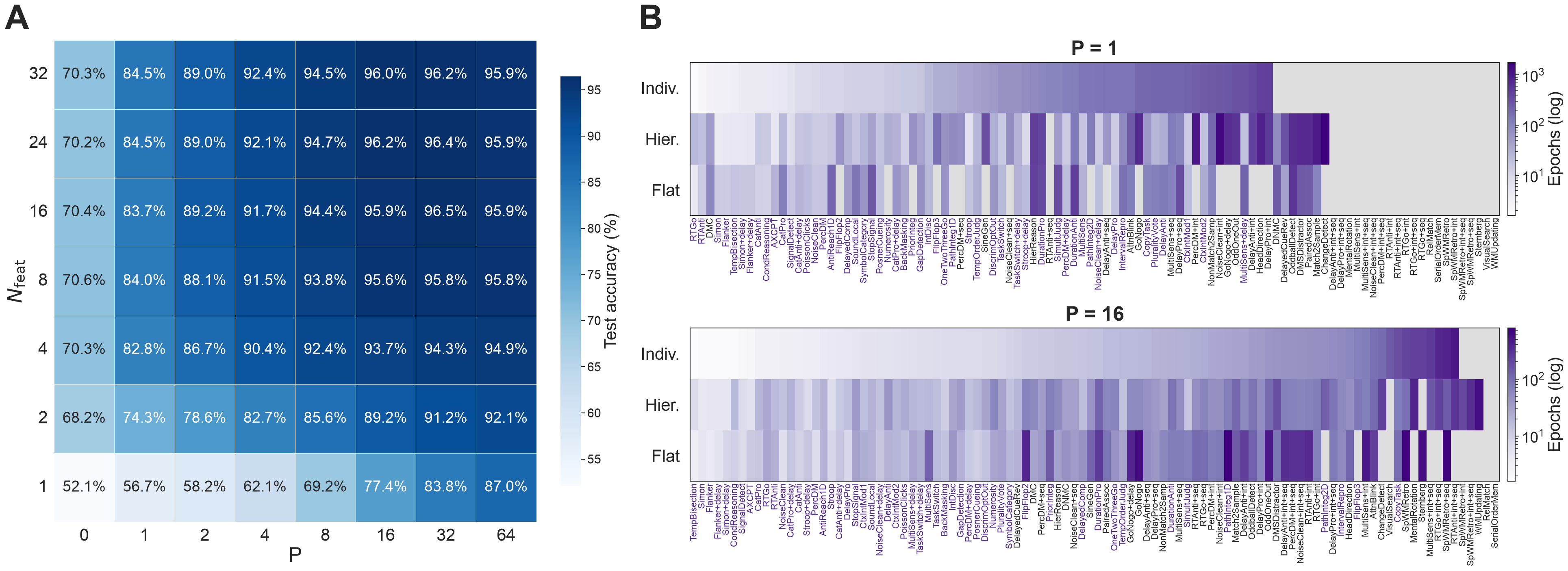}
    \caption{\textbf{A}: Test accuracy of the hierarchical model (AL-RNN, $M = 64$) across task-embedding dimension $N_\text{feat}$ and nonlinear units $P$, averaged across 5 seeds. \textbf{B}: Epochs to reach $>80\%$ test accuracy per task at $P = 1$ and $P = 16$ for individual models ($M = 64$), the hierarchical model ($N_\text{feat} = 16$) and a parameter-matched flat model ($M = 234$), averaged over the seeds that reached threshold. Grey: never reached. Purple names: tasks solved by a fully linear individual model.}
    \label{fig:performance_complexity}
\end{figure}

\paragraph{Individual task complexity varies across the battery} 
To assess how model complexity affects joint performance, we varied the number of nonlinear units and the dimensionality of the task features, including a comparison to fully linear RNNs. Performance in the hierarchical model improved smoothly along both dimensions (Fig.~\ref{fig:performance_complexity}\textbf{A}), but relatively little capacity was needed to approach the performance ceiling. With only $N_{\mathrm{feat}}=4$ task features, the model reached approximately $95\%$ accuracy at sufficient nonlinear capacity, while introducing just two nonlinear units increased performance from roughly $70\%$ to $89\%$ with sufficient hierarchical features. This high performance of simple models is striking given the apparently demanding cognitive and computational requirements represented in the battery, and only a small subset of tasks requires substantial model complexity. For instance, linear hierarchical and individually trained models already solve 45 and 53 tasks respectively, and a small number of nonlinear units suffices to solve most other tasks. This extends the observation of \citet{khona2023winninglotteryneuralconnectivity} that commonly used cognitive tasks often admit surprisingly simple recurrent solutions. Other tasks, like the classic neuroscientific \textit{head direction integration} paradigm, require nonlinear dynamics to implement more complex attractor structures (Fig. \ref{fig:head_discriminator_task}).
%Of the 14 tasks shared between their battery and ours, 10 are solved by linear models, while the remaining four require at most 2 out of 64 nonlinear units.
We found that linear solvability was also a proxy for how fast a task was learned. We measured the time required for each task to reach $>80\%$ test accuracy and compared it with the task's performance in an individually trained linear model. Tasks well solved by linear models converged substantially faster than tasks requiring nonlinearity. For example, in the joint model with $P=16$ nonlinear units, the tasks classified as individually linear were learned approximately four to five times faster than the remaining tasks (Fig. \ref{fig:performance_complexity}\textbf{B}). This relationship was observed for both individually trained models and the hierarchical joint model, whose convergence times were highly consistent across tasks (Fig. \ref{fig:performance_complexity}\textbf{B}, Spearman $\rho\approx0.86$). The flat joint model instead exhibited several pronounced task-specific outliers, indicating that the unstructured shared parameterization introduces additional optimization difficulty for some tasks, particularly without sufficient nonlinear capacity.

\subsection{Mechanisms for Task Separation}
A shared recurrent model can separate tasks in at least three ways: by assigning them different task-specific features, by routing them through different linear regions, or by embedding their computations in different geometric subspaces of the state space. The model comparisons in Table~\ref{tab:main_performance} allow us to examine these mechanisms separately. The flat linear model has access only to geometric separation and solves just 14 tasks, although 52 tasks are individually linearly solvable. Adding hierarchical task features recovers the individually linear solutions, but cannot solve tasks that require nonlinearity in isolation. Conversely, adding nonlinear units to the flat model both recovers individually linear tasks and enables some intrinsically nonlinear tasks, reaching 65 solved tasks at its best-performing setting. However, performance peaks with sparse nonlinearity (16/234 nonlinear units), and declines as further nonlinear units are added (Fig. \ref{fig:samples_vs_nonlinearity}, mirroring similar findings by \citealp{brenner_uncovering_2026}). The hierarchical nonlinear model, which combines all three forms of organization, performs best, while retaining an individual DS model per task that can be analyzed in terms of its dynamical mechanism.

\paragraph{Individual vs. joint solvability}
We next tested the relationship between individual and joint complexity explicitly by combining sets of tasks that were each individually linearly solvable. As the number of jointly trained tasks increased, more nonlinear units were required to solve the complete set (Fig.~\ref{fig:nonlinear_joint}). The AL-RNN lets us examine how nonlinear separation is implemented. Let $R$ denote the active piecewise-linear region, observed through its bitcode $\bm{b}_t$ at time $t$, and let $Q$ denote task identity. The conditional entropy $H(R\mid Q)$ measures variation in region usage within tasks, while the mutual information $I(Q;R)=H(R)-H(R\mid Q)$ measures how much task identity can be inferred from the active region alone. High $I(Q;R)$ therefore indicates that tasks occupy distinct linear subregions, whereas low $I(Q;R)$ indicates larger overlap. In both model classes, $I(Q;R)$ increased with nonlinear capacity. However, $I(Q;R)$ was generally much larger in the flat model than in the hierarchical one (e.g. for $P=16$, $1.6$ in the hierarchical model but approximately $5.0$ bits in the flat model, close to the task entropy ceiling of $\approx6.64$ bits). The flat model therefore relies more heavily on leveraging different linear regions to separate tasks, whereas the hierarchical model can implement part of this separation through its task features. We find a similar pattern in recurrent-state geometry. Task-specific overlap of the first five principal components (PCs) is larger in the hierarchical than in the flat model (approximately $0.26$ versus $0.14$, Fig. ~\ref{fig:jointflat_var_overlap}), indicating greater sharing of task subspaces. At the same time, the first five PCs of the global state space explain more variance in the hierarchical model ($0.43$ versus $0.32$), consistent with a more compact overall geometry. Thus, the flat model separates tasks more strongly in both linear region usage and state space geometry, whereas the hierarchical model can additionally leverage the flexibility of the task features to enable sharing and separation more effectively.

\subsection{Task Similarity Spaces}
\begin{figure}[!htb]
    \centering
    \includegraphics[width=0.99\linewidth]{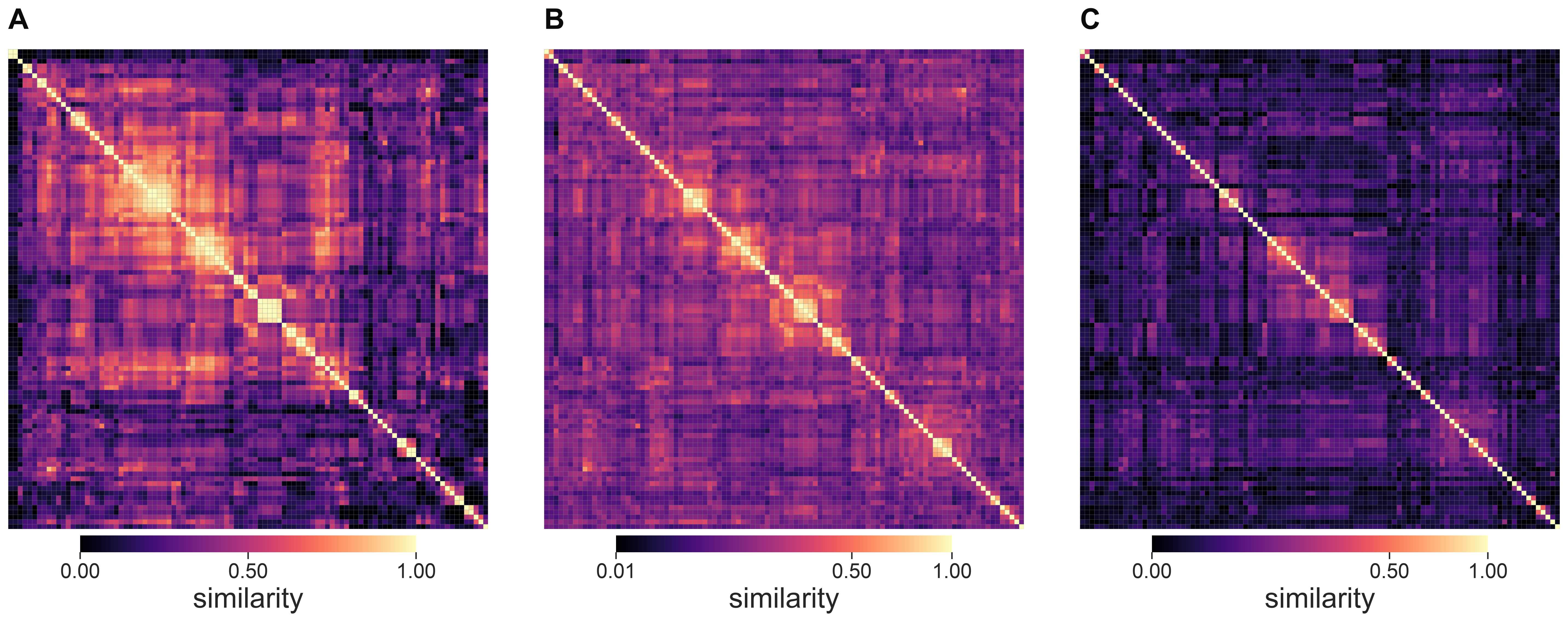}
    \caption{Task-similarity spaces learned by the hierarchical AL-RNN for \textbf{A} task features, \textbf{B} geometry, and \textbf{C} linear region usage, for $N_{\mathrm{feat}}=16$ and $P=16$. Tasks are ordered by clustering the hierarchical task features (clustering details and task labels reported in Appx. \ref{appx:clustering}).}
    \label{fig:task_spaces}
\end{figure}

The preceding results identify three mechanisms through which tasks can be organized within the shared model: hierarchical task features, linear region usage, and the geometric overlap of task-specific activity subspaces. Each mechanism defines a pairwise similarity measure between tasks (cosine similarity of task features, overlap of linear region distributions, and overlap of the first five activity PCs) which we interpret as complementary \textit{task-similarity spaces} (Fig.~\ref{fig:task_spaces}). We next investigate their robustness, agreement, and relationship to transfer and interference.

We first assessed reproducibility by correlating the off-diagonal entries of the $100\times100$ task-similarity matrices across random seeds. All three AL-RNN spaces were consistent across seeds ($r\approx0.76$-$0.78$ for $N_\text{feat}=16, P=16$; Table~\ref{tab:task_space_correlations}), with similar values in LSTMs and GRUs (Table \ref{tab:cross_architecture_task_spaces}). Agreement across the different AL-RNN spaces was lower ($r=0.47$-$0.70$), indicating that they capture related but non-identical aspects of task organization. Feature-similarity structure also generalized moderately across architectures (AL-RNN--LSTM: $r=0.53$; AL-RNN--GRU: $r=0.57$). Thus, task organization depends partly on the recurrent architecture, but substantial structure remains reproducible across both initializations and model classes.

\begin{figure}[htb!]
    \centering
    \includegraphics[width=.99\textwidth]{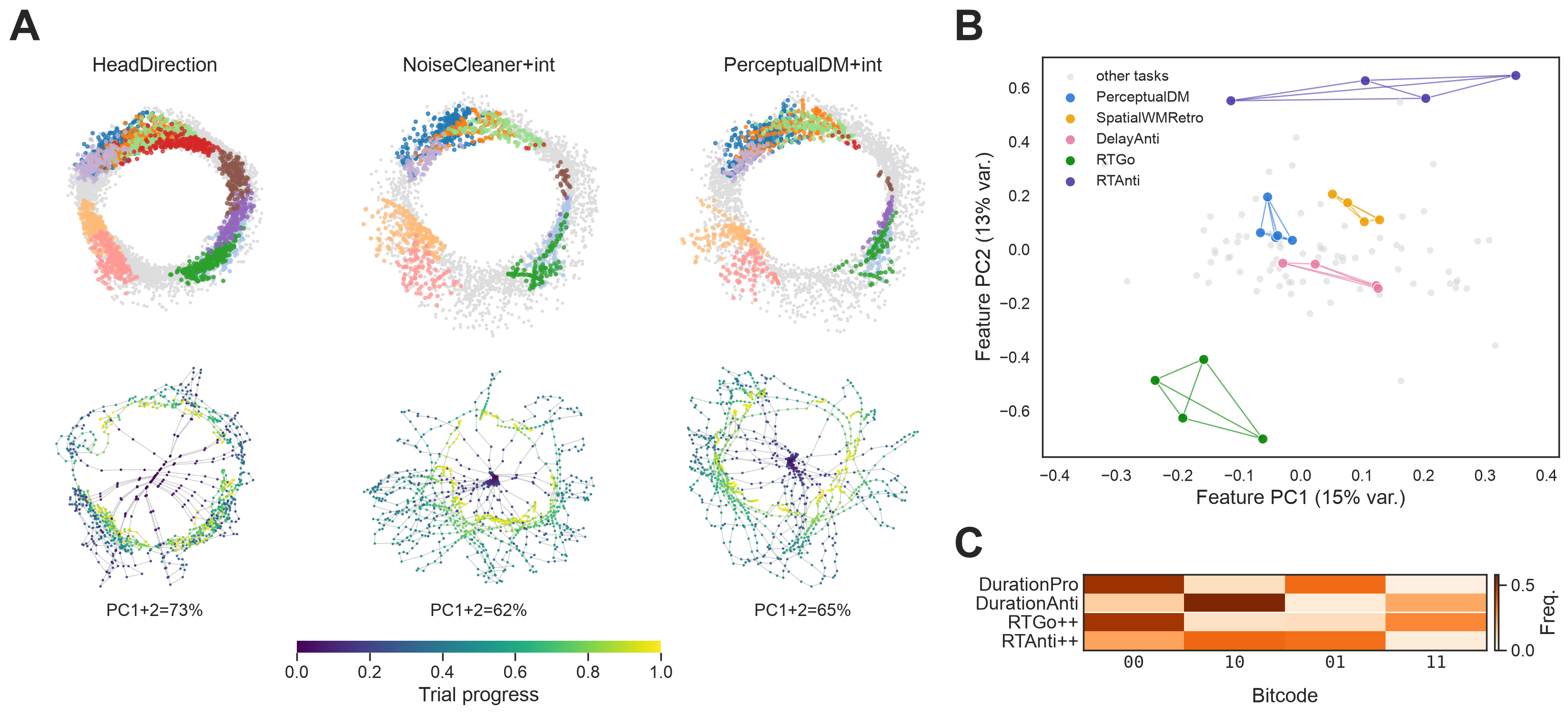}
    \caption{\textbf{A}: Example of three tasks (HeadDirection, NoiseCleaner+int and PerceptualDM+int) close in the task similarity spaces ($P=16, N_\text{feat}=16$) leveraging shared attractor structure. Top: Network states during the integration/delay phase, projected onto the first two PCs of HeadDirection's state space. The ring attractor is approximated by piecewise linear subregions (see also Fig. \ref{fig:head_discriminator_task}). Points of the 12 most visited regions, coloured by the linear region they occupy (grey for all others), appear at similar positions on all three rings. Bottom: 30 example trials per task, coloured by trial progress (dark = trial start, yellow = trial end). \textbf{B}: Five tasks with their respective task modifiers in the hierarchical feature space cluster together. \textbf{C}: Pro and Anti variants of the same tasks are placed in opposing linear subregions. ++ denotes +int+seq modifier.}
    \label{fig:trajectories}
\end{figure}

\paragraph{Task spaces reveal interpretable structure} 
The three task spaces can be queried along different dimensions to examine how the model organizes task solutions. We provide three illustrative, non-exhaustive examples. Task geometry similarity reveals that tasks with related dynamical computations occupy visually similar geometric organizations, including recurring rotations, trajectories, and other low-dimensional motifs (Fig.~\ref{fig:similar_trajectories}). For example, HeadDirection and its nearest neighbours in both feature and geometric space, and among the closest tasks in linear region usage, NoiseCleaner$+$int and PerceptualDM$+$int, reuse a very similar piecewise-linear approximation to a ring attractor (Fig.~\ref{fig:trajectories}\textbf{A}, see also Fig.~\ref{fig:head_discriminator_task}). Thus, proximity in the learned task spaces can reflect shared dynamical objects that are repurposed across tasks. Second, compositional variants of the same tasks cluster in the hierarchical feature space. For the 49 tasks belonging to 15 base families with at least one modified counterpart, the mean cosine similarity distance to variants was $0.21 \pm 0.02$, while the mean distance to other tasks was almost three-fold, $0.62 \pm 0.03$, $p<.001$ for all seeds (Fig. \ref{fig:trajectories}\textbf{B}). Finally, pro-anti pairs of the same tasks are often distributed along different linear subregions, in line with the observation that the model requires nonlinearity to jointly solve similar tasks with opposing response rules (Fig. \ref{fig:trajectories}\textbf{C}, see also Fig. \ref{fig:pro_anti_bitcodes}).

\subsection{Task representations predict transfer and adaptation}
\paragraph{Task isolation predicts individual-model transfer} We then evaluated transfer between individually trained models by initializing each target task from every other task's trained solution and measuring early learning over the first 100 epochs (see Appx.~\ref{app:pairwise_transfer}). Pretraining improved learning relative to random initialization in approximately $70\%$ of source-target pairs, but several targets were learned more effectively from scratch than from most pretrained sources. Thus, pretrained task solutions provide useful priors on average, but can act as negative priors for particular tasks. We then asked whether this source quality was related to the task similarity spaces learned by the joint model ($P=16$, $N_\text{feat}=16$). Across tasks, more isolated representations were modestly associated with poorer transfer sources in both task geometry ($\rho=-0.26$, $p=0.01$) and linear region usage ($\rho=-0.23$, $p=0.02$, also see Fig. \ref{fig:negative_transfer}). These effects do not fully explain transfer, but provide preliminary evidence that task organization in the joint model reflects aspects of transferability between independently trained task solutions.

\paragraph{Leave-one-out adaptation}
We finally tested whether a model pretrained on 99 tasks could learn to solve the remaining task by adapting only a task-specific feature vector ($N_\text{feat}=16$). For each held-out task, we froze the shared parameters and initialized the feature vector either from the mean of the 99 pretrained task features or from the best-performing source vector among them. This procedure preserves the pretrained tasks by construction and tests whether the shared backbone already contains a suitable solution that can be accessed through its low-dimensional task interface. Adaptation reached $92.1\% \pm 0.1\%$ accuracy when initialized from the best source vector, close to the $96\%$ achieved by the fully trained joint model, and even slightly outperforming individually trained full models ($\approx 91\%$, Table \ref{tab:main_performance}). In contrast, initialization from the mean feature vector reached only $84.5\%\pm 1.9\%$. The best-source result should be viewed as an oracle upper bound and not a realistic initialization strategy, but it shows that the pretrained backbone in principle already supports near-best performance, even though only the $16$ task-feature parameters need to be adapted, rather than the roughly $800$ parameters of an equivalent individually trained task model. However, the gap from mean initialization further shows that this solution is not generally recovered by feature optimization from an arbitrary point, including across all learning rates $\eta\in\{10^{-4},10^{-3},5\times10^{-3},10^{-2},5\times10^{-2}\}$ tested, indicating a non-smooth loss landscape. 

Since the same procedure (initializing fine-tuning from another task's feature vector) can be repeated for every source-target pair, it also provides a measure of how useful each task representation is as an initialization, where we quantified training benefit using the same measure from the previous section (Appx.~\ref{app:pairwise_transfer}). We then related source quality to task isolation, defined as the mean distance of a task from the other tasks in each similarity space. Source quality was negatively correlated with isolation in all three spaces: $\rho=-0.74$ for task geometry, $\rho=-0.86$ for task features, and $\rho=-0.53$ for linear region usage, with $p<.001$ in all cases. The relationship in the hierarchical feature space is partly expected because adaptation is performed by optimizing coordinates in that space. However, the corresponding associations with independently measured geometry and linear region usage similarity show that task positions learned during joint training more generally carry information about the functional reusability of those positions for adapting to other tasks. Thus, the same pre-training and adaptation framework both supports learning new tasks without forgetting and indicates which task representations are useful for successful transfer learning.

\section{Discussion}
We trained a hierarchical AL-RNN jointly on 100 heterogeneous cognitive tasks, using it to study how task computations interact within a shared dynamical backbone. We identified different mechanisms for separating and combining tasks through hierarchical task features, linear region usage, and task geometry. Together, these mechanisms allow the model to reuse shared structure while avoiding negative interference. The resulting task spaces provide several reproducible but distinct measures of task similarity. This offers a representation-based alternative to relying on task names or verbal taxonomies, which may obscure the computational relationships between tasks, and a common interface for comparing task representations across models.

Our results also highlight challenges that become especially important as multitask recurrent models are scaled. The framework reveals substantial heterogeneity in individual task complexity, particularly in the amount of nonlinear capacity required for learning, even when tasks appear similarly demanding from their verbal descriptions. Individual tasks can be simple while their joint implementation requires additional model complexity, and increasing the number and diversity of tasks can increase the need for such capacity. Transfer is consequently not uniformly positive, with some tasks systematically interfering with others. Task isolation in the derived task similarity spaces relates to negative transfer in pairwise pre-training. We further find that task-conditioned parameterization provides an efficient adaptation mechanism, where new tasks can be learned by modifying only a low-dimensional feature vector, avoiding catastrophic forgetting by construction. This suggests a route toward reusable dynamical backbones for cognitive computation, although the loss landscape for this adaptation is sufficiently irregular that success depends strongly on initialization. The results on negative transfer present an interesting avenue for comparing model-predicted task interactions with neural data: for example, do isolated tasks in the model also predict slower adaptation when humans or animals learn them sequentially?

To facilitate adaptation and identify compatible task pairings, future work could learn an encoder that provides informative priors for task-feature initialization \citep{vermani_meta-dynamical_2024}, building on task-embedding methods such as Task2Vec \citep{achille_task2vec_2019}. Task features could also be organized according to dynamical properties \citep{trede_topological_2026}, potentially improving both adaptation and interpretability. Finally, comparing the learned task representations with complementary characterizations based directly on dynamics may clarify which notions of task similarity generalize across methods \citep{ostrow_beyond_2023, huang_inputdsa_2026}.

A limitation of the present study is that its conclusions are based on a particular task battery and model class, while the field still lacks standardized benchmarks for evaluating transfer and interference across cognitive tasks. Moreover, the same task may admit multiple dynamical solutions \citep{qian_discovering_2025, tolmachev_single-unit_2025}, and the representations identified here may depend on architecture and optimization. We therefore view the task suite, evaluation, and pairwise transfer protocols introduced here as a contribution towards a reproducible basis for comparing multitask models. Extending these comparisons across architectures, learning rules, and task batteries will be necessary to determine which findings generalize.

\section*{Reproducibility Statement}
The code for this paper is available at 
\url{https://github.com/psynamic-lab/multitask_interactions.git}.

\section*{AI use statement}

In this work, we used generative AI tools to assist with the initial generation and implementation of candidate task paradigms. Candidate tasks, their literature sources, and their final implementations were reviewed and verified by the authors. The model architecture and training framework were built on existing implementations from the literature, and AI assistance was used primarily for processing results and implementing visualizations. For writing, generative AI was used for editing and phrasing suggestions based on author-provided structure and content. We reviewed all AI-assisted code, analyses, figures, and text, and take full responsibility for the final content, claims, and artifacts of this work.

\section*{Acknowledgements}
This work was supported by the DYNAMIC Centre, funded by the LOEWE program of the Hessian Ministry of Science and Arts, grant LOEWE 1/16/519/03/09.001(0009)/98 (to MB), and by the Ernst Strüngmann Foundation.

%\newpage

\bibliography{bibliography,task_references}

\bibliographystyle{plainnat}

\newpage
\appendix

\section{Methodological Details}

\subsection{Task Suite} \label{app:task_suite}

We train on a battery of 100 supervised cognitive tasks spanning working memory, decision-making, perception, attention, timing, reasoning, and motor response. The battery comprises $66$ base task identities and $34$ compositional variants generated from selected bases using four modifiers. Each modified variant is treated as a separate task with its own task embedding.

All tasks share a common trial interface. Inputs are zero-padded to a common input dimension, and targets are represented in a fixed five-channel readout comprising a fixation channel and task-dependent behavioural channels. Trials may contain context, stimulus, delay, and response phases, with phase durations resampled across trials.

Table~\ref{tab:tasks} lists the $66$ base task identities and their readout types. The modifiers are defined in Sec.~\ref{modifiers}. Here, \emph{angular} denotes a directional output represented by $(\cos\theta,\sin\theta)$, \emph{binary} a thresholded decision, \emph{scalar} a one-dimensional continuous output, \emph{vector} multiple continuous coordinates, \emph{categorical} a choice among multiple alternatives, and \emph{bits} the signs of multiple binary outputs. A star ($^\star$) marks outputs scored at every response timestep.

Because the task battery contains tasks with different temporal structures and readouts, we use a common masked-MSE objective for joint training, with task-specific masks selecting the behaviourally relevant timesteps and readout channels, including the fixation channel:

\begin{equation} \label{eq:loss}
\mathcal{L} \;=\; \frac{1}{B}\sum_{b=1}^{B} \frac{1}{L_{k(b)}^{(0)}}\,
\frac{1}{|M_b|\,|C_b|}\sum_{t\in M_b}\sum_{c\in C_b}\bigl(\hat y_{b,t,c}-y_{b,t,c}\bigr)^2,
\end{equation}

where $M_b$ denotes the scored time steps, $C_b$ the scored output channels for trial $b$, $k(b)$ is the task of trial $b$ and $L_k^{(0)}$ is the mean of the unweighted per-trial loss (the inner term of Eq. \ref{eq:loss}) over task $k$'s training trials at initialization, computed once before training and held fixed. Normalizing each trial by the number of scored time-channel entries prevents tasks with longer sequences or more output channels from dominating the joint gradient, and dividing by $L_k^{(0)}$ expresses each task's loss relative to its value at initialization, so that tasks whose targets differ in magnitude contribute on a comparable scale. We use no task-specific loss function.

We report accuracy separately from the training loss using a task-specific, trial-level success criterion. For single-response tasks, a trial is counted as correct if the final scored output lies within the prescribed tolerance of the target. Reaction-time tasks are evaluated at the response timestep following stimulus onset. For sequential tasks, accuracy is the fraction of scored timesteps satisfying the corresponding criterion. Following \cite{yang_task_2019}, we set the angular threshold to $36^\circ$. To make scores comparable across response types, we use angular tolerances for directional outputs and a scalar/vector tolerance set to $0.35\,\sigma$ of the task's target distribution. This mapping between output tolerances and trial success is necessarily a modeling assumption, so we assess the robustness of this choice by recomputing battery accuracy across a range of angular thresholds (Table~\ref{tab:battery_accuracy}). Although the absolute accuracy changes and the chosen $36^\circ$ criterion produces a mild ceiling effect, the relative ordering of model capacities is preserved across thresholds. In addition to mean accuracy across tasks, we report the number of solved tasks, defining a task as solved when its accuracy $>80\%$.

\input{tables/table_accuracy}

\subsection{Model Training}
\label{app:training_procedure}

For each task, we used 200 training trials and 256 evaluation trials per task, with balanced task indices ensuring that every task contributed equally. Trials were sampled in batches of 64 and each epoch traversed the complete training pool once without replacement. Since trials differed in length and channel dimensionality, batches were zero-padded to their maximum dimensions and accompanied by timestep masks, so padding did not contribute to the masked-MSE objective defined above.

The principal model-capacity hyperparameters were the latent dimension $M$, the number of nonlinear units $P$, and the task-feature dimension $N_{\mathrm{feat}}$. We varied $P$ and $N_{\mathrm{feat}}$ in the main capacity experiments. In preliminary experiments, performance largely plateaued at $M=64$, and larger latent spaces yielded similar accuracy. We therefore fixed $M=64$ for the main experiments. Initialization and optimization settings were chosen following the AL-RNN training procedure of \citet{brenner_learning_2024}.

For the hierarchical models, task features and task-specific noise covariances were optimized jointly with the shared parameters using Adam \citep{kingma_adam_2017}. We implemented separate parameter groups for task-specific and shared parameters, with a common learning rate of $5\times10^{-4}$ and weight decay of $10^{-4}$. The shared recurrent projection was initialized with Xavier gain $0.05$.

For training the AL-RNN, the objective also included the manifold attractor regularization (MAR, \citealp{schmidt_identifying_2020}) with ($\tau=0.01$), applied to the first half of the recurrent units. This regularization encourages self-recurrent weights toward one and off-diagonal recurrent weights and biases toward zero, implementing slow time scales commonly useful in computational tasks \citep{schmidt_identifying_2020}, and incorporated as an architectural bias into LSTMs and GRUs. The total objective was Eq. \ref{eq:loss} plus this regularization term. Gradients were clipped to a global norm of 10, and updates were skipped if the post-clipping norm was non-finite.

Learning rates were linearly warmed up over 100 epochs and then exponentially decayed by a factor of $0.999$ per epoch. If a non-finite loss or gradient was detected, the learning-rate scale was halved for the remainder of the run. We evaluated the complete validation set after every epoch and tracked total loss and per-task accuracy. The checkpoint with the lowest validation loss was retained. After the first 300 epochs, we used early stopping with a patience of 300 epochs without improvement, training for up to 3000 epochs in total. All reported accuracies were scored on an independent test set with 256 samples per task.

For the leave-one-out experiments, we trained one hierarchical model per held-out task on the
remaining 99 tasks ($M = 64$, $P = 16$, $N_\text{feat} = 16$), using the same data, optimization
settings and stopping criteria as above, for three seeds each (300 models in total). To adapt a
model to its held-out task, we froze all shared parameters and optimized only the new task's
feature vector, with a learning rate of $10^{-2}$ and a batch size of 64 on 200 training trials,
for up to 700 epochs with early stopping, evaluating on 128 test trials per epoch.

The flat joint models consisted of a single AL-RNN with one set of shared parameters trained on all
100 tasks. Task identity was provided as a one-hot vector appended to the input. They were trained
with the same optimization settings as the hierarchical models, except that 128 test trials per task were used.

Individual models were trained separately on each task as flat AL-RNNs with a single set of
shared parameters and no task input ($M = 64$, $P \in \{0,1,2,4,8,16\}$, 20, 50 or 200 training
trials, five seeds). They used the same learning rate, weight decay, gradient clipping and MAR
regularization as the hierarchical models, no learning-rate warm-up and up to 500 epochs. Early stopping was used with a patience of 150 epochs. For the pairwise transfer experiments, each trained source model was used to initialize all parameters of a model trained on each of the other 99 targets, using the same settings as the individual models, for five seeds.

\paragraph{Clustering Algorithm} \label{appx:clustering}

\begin{figure}[htb!]
    \centering
    \includegraphics[width=.99\textwidth]{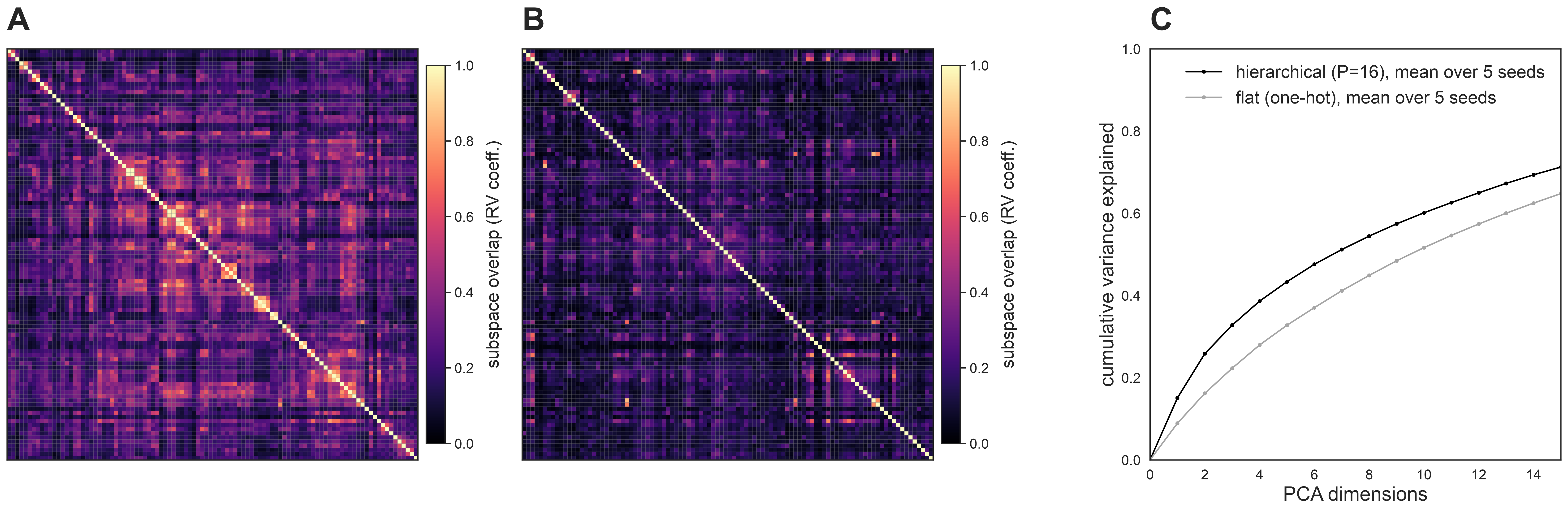}
    \caption{Geometric task similarity matrices for the hierarchical (\textbf{A}) and flat  models (\textbf{B}) with $M=64$ latent dimensions and the same task order in both. \textbf{C}: Cumulative variance explained via the first 15 principal components. The hierarchical model compresses activity into fewer dimensions, retaining substantially more overlap among tasks.}
    \label{fig:jointflat_var_overlap}
\end{figure}

For the results in Figs. \ref{fig:task_spaces} and \ref{fig:jointflat_var_overlap}, we leveraged a joint hierarchical clustering across all three similarity spaces. Tasks were ordered by agglomerative hierarchical clustering with average linkage (UPGMA, \citealp{sokal1958statistical}) on the dissimilarity $1-s$, where s is the mean of the three seed-averaged similarity matrices, with each min-max normalized on its off-diagonal entries. The leaves were then reordered with optimal leaf ordering \citep{barjoseph2001fast}, using the default implementation in SciPy \citep{virtanen2020scipy}. The resulting clustering-based task order used in Figs.~\ref{fig:task_spaces} and \ref{fig:jointflat_var_overlap} is: 
1.\ FlipFlop2; 2.\ FlipFlop3; 3.\ CopyTask; 4.\ GoNogo; 5.\ GoNogo+delay;
6.\ HierReason; 7.\ TaskSwitch; 8.\ TaskSwitch+delay; 9.\ GapDetection;
10.\ DelayedComparison; 11.\ IntDisc; 12.\ DurationAnti; 13.\ DurationPro;
14.\ StopSignal; 15.\ CatPro; 16.\ CatPro+delay; 17.\ DMC; 18.\ DMSDistractor;
19.\ Match2Sample; 20.\ NonMatch2Sample; 21.\ CatAnti+delay; 22.\ CatAnti;
23.\ SymbolCategory; 24.\ SimultaneityJudgment; 25.\ TemporalOrderJudgment;
26.\ BackwardMasking; 27.\ DelayedCueReversal; 28.\ Stroop+delay; 29.\ Stroop;
30.\ Flanker; 31.\ Simon; 32.\ Flanker+delay; 33.\ Simon+delay; 34.\ AXCPT;
35.\ ConditionalReasoning; 36.\ DNMC; 37.\ PairedAssoc; 38.\ CtxIntMod1;
39.\ DelayPro; 40.\ PerceptualDM; 41.\ NoiseCleaner; 42.\ NoiseCleaner+delay;
43.\ PerceptualDM+delay; 44.\ MultiSens; 45.\ MultiSens+delay; 46.\ CtxIntMod2;
47.\ PriorIntegration; 48.\ NoiseCleaner+seq; 49.\ PerceptualDM+seq;
50.\ MultiSens+seq; 51.\ MultiSens+int+seq; 52.\ MultiSens+int;
53.\ PerceptualDM+int+seq; 54.\ NoiseCleaner+int+seq; 55.\ NoiseCleaner+int;
56.\ PerceptualDM+int; 57.\ HeadDirection; 58.\ DelayPro+seq;
59.\ DelayPro+int+seq; 60.\ DelayPro+int; 61.\ DelayAnti+int;
62.\ DelayAnti+int+seq; 63.\ DelayAnti+seq; 64.\ DelayAnti;
65.\ IntervalReproduction; 66.\ OneTwoThreeGo; 67.\ Numerosity;
68.\ AntiReach1D; 69.\ SoundLocalization; 70.\ PoissonClicks;
71.\ PluralityVote; 72.\ PathIntegration1D; 73.\ PathIntegration2D;
74.\ PosnerCueing; 75.\ WMUpdating; 76.\ RelationalMatch; 77.\ MentalRotation;
78.\ Sternberg; 79.\ ChangeDetect; 80.\ SerialOrderMemory; 81.\ VisualSearch;
82.\ SpatialWMRetro+int+seq; 83.\ SpatialWMRetro+int; 84.\ SpatialWMRetro;
85.\ SpatialWMRetro+seq; 86.\ DiscriminationOptOut; 87.\ SignalDetection;
88.\ TemporalBisection; 89.\ OddballDetection; 90.\ SineGen;
91.\ AttentionalBlink; 92.\ RTAnti; 93.\ RTAnti+seq; 94.\ RTAnti+int+seq;
95.\ RTAnti+int; 96.\ RTGo+int; 97.\ RTGo+int+seq; 98.\ RTGo+seq;
99.\ RTGo; 100.\ OddOneOut.
%\end{quote}
%\normalsize

Although the three similarity spaces show moderate pairwise agreement, this shared structure does not support a single robust partition of the full task battery. Clustering the spaces independently yielded only weak agreement across cluster numbers ($\mathrm{ARI}\approx0.1$-$0.35$ for $k=2$-$20$). This partly reflects their different similarity distributions: feature similarity has higher and more structured off-diagonal values than geometry or linear region usage (see Fig. \ref{fig:task_spaces}, and consequently contributes more strongly to the joint clustering. Moreover, silhouette analyses did not identify a stable preferred number of clusters. Even in the feature space, the highest scores were obtained by separating a small number of outlier tasks rather than by revealing a meaningful multiway partition. Nevertheless, targeted comparisons, such as the proximity of task variants to their base tasks, recover expected local structure, indicating that the spaces capture meaningful task relationships despite the absence of a single globally robust clustering.

\paragraph{Pairwise transfer between individually trained models}
\label{app:pairwise_transfer}

To assess transfer independently of the shared multitask backbone, we evaluated all ordered pairs of individually trained task models. For each source task $A$ and target task $B$, we first trained a model on $A$ to convergence and then used its parameters to initialize training on $B$. We compared these source-initialized runs with training on $B$ from the same random initialization distribution. All target runs used the same architecture, training data, evaluation data, and optimization schedule; only the source initialization differed.
We quantified early learning using the negative mean log test loss over epochs $6\text{--}100$,
\begin{equation}
S(A\!\rightarrow\!B)
=
-\frac{1}{95}\sum_{e=6}^{100}
\log \mathcal{L}_{\mathrm{test}}^{(e)}(A\!\rightarrow\!B),
\end{equation}
such that larger values indicate lower test loss and more effective early adaptation. The first five epochs were excluded because they contained large initialization-dependent transients. For each target $B$, we standardized the 99 source-specific scores, $z(A\!\rightarrow\!B)=\frac{S(A\!\rightarrow\!B)-\mu_B}{\sigma_B}$, where $\mu_B$ and $\sigma_B$ are the mean and standard deviation of $S(A\!\rightarrow\!B)$ across all 99 source tasks. Scores were computed for five random seeds and aggregated using the median across seeds. We define the \emph{source quality} of task $A$ as the mean of $z(A\!\rightarrow\!B)$ across all target tasks $B\neq A$.

We additionally evaluated training from random initialization as a reference source. Pretrained task models improved early learning relative to random initialization in approximately $70\%$ of source--target pairs. However, this benefit was task-dependent: for some targets, training from scratch outperformed initialization from most source tasks. Thus, transfer was beneficial on average but could act as a negative prior for particular target tasks.

\begin{figure}[htb!]
    \centering
    \includegraphics[width=.90\textwidth]{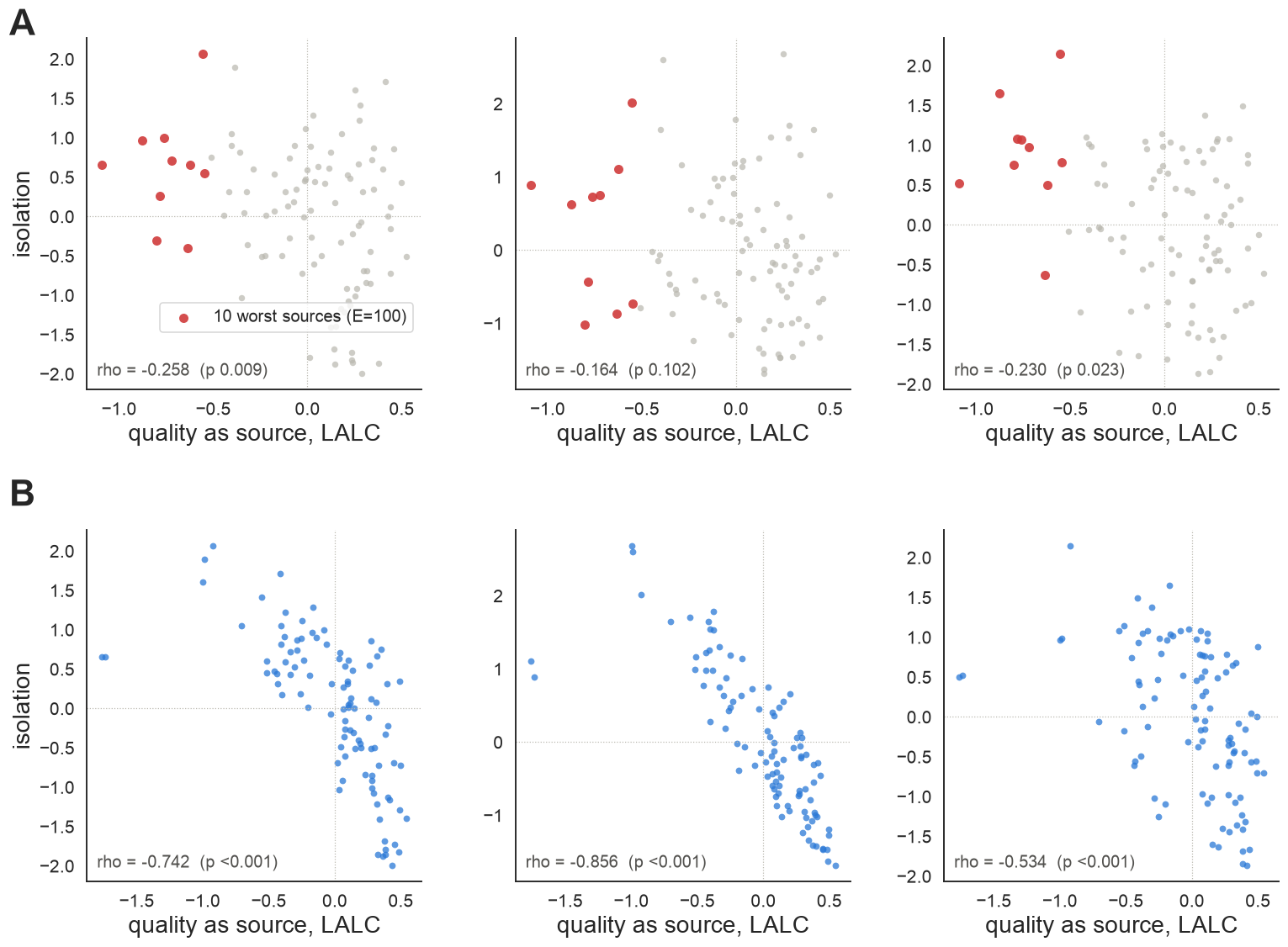}
    \caption{Isolation in the hierarchical AL-RNN ($N_\text{feat}=16, P=16$) feature spaces (geometry, task features and linear region usage from left to right), plotted against two different measures of source quality. \textbf{A}: Source quality is computed from the individual-model transfer matrix: one network per task ($P=16$), fine-tuned onto every other task. Red points are the 10 worst sources when considering the first 100 training epochs (E100). \textbf{B}: Source quality computed from the leave-one-out experiments, where the task feature for the target task was fine-tuned initialized from the source task's task feature.}
    \label{fig:negative_transfer}
\end{figure}

To relate this result to the representations learned in the joint model, we quantified task isolation separately in each task-similarity space. For a task $A$, isolation was defined as its mean distance to all other tasks, standardized across the 100 tasks. We then computed Spearman correlations between source quality and isolation. Source quality was negatively associated with isolation in recurrent-state geometry ($\rho=-0.26$, $p=0.01$) and nonlinear-region usage ($\rho=-0.23$, $p=0.02$), indicating that tasks occupying more isolated representations in the joint model tended to provide less useful initializations for independently trained task models (see Fig. \ref{fig:negative_transfer}). These relationships were modest and do not fully explain pairwise transfer, but indicate that joint-model task organization captures some aspects of transferability.

\input{tables/table_accuracies}

\newpage

\section{Further Results}

\begin{figure}[htbp]
    \centering
    \includegraphics[width=0.75\textwidth]{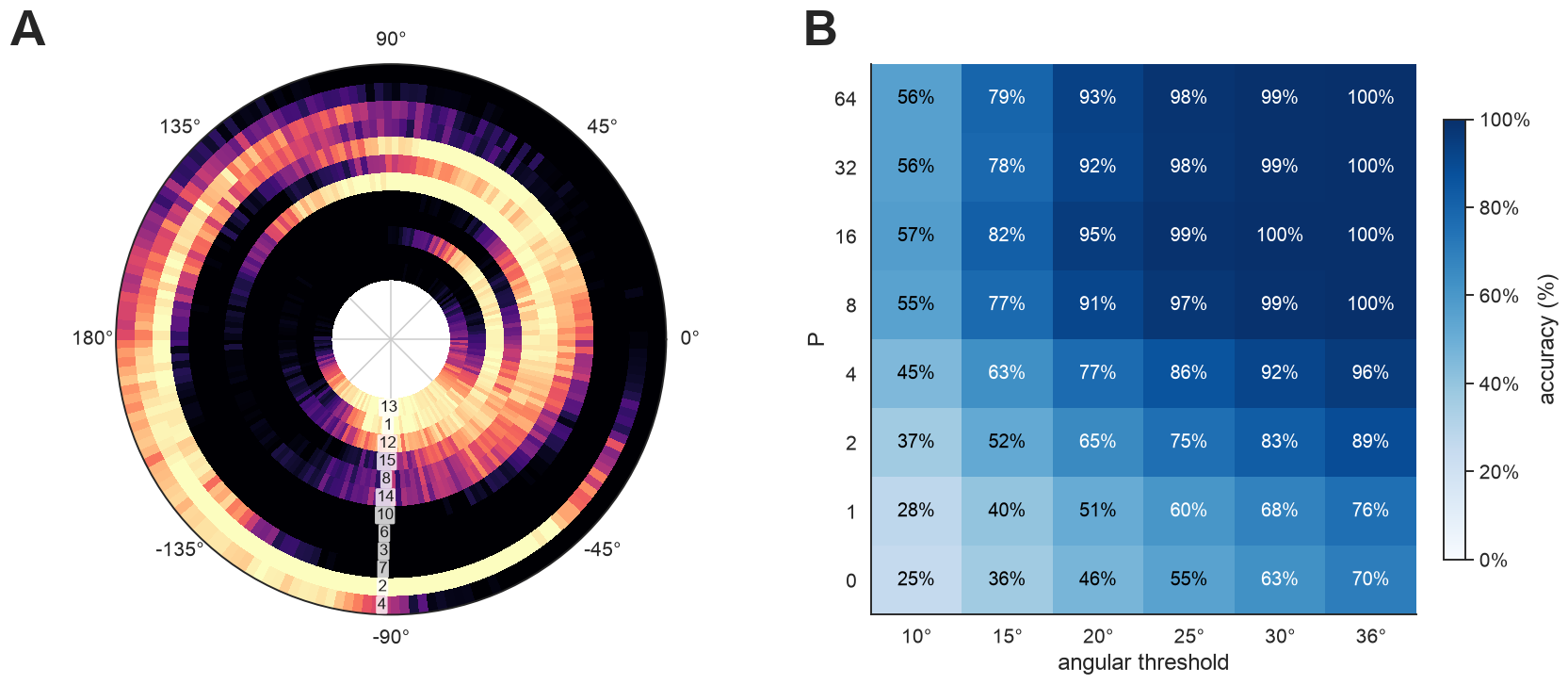}
    \caption{Hierarchical AL-RNN ($P=16, N_\text{feat}=16$) trained on the HeadDirection task. \textbf{A}: Probability that each ReLU unit is active, $P(\text{unit}>0)$, as a function of heading (one concentric ring per unit, ordered by the onset of its active arc). Each unit switches on over a single contiguous arc, partitioning the circle into sectors whose boundaries tile the ring attractor (Fig.~\ref{fig:trajectories}). An exactly continuous attractor would require a one-dimensional manifold of fixed points with marginal stability ($|\lambda|=1$) everywhere along it. This is a measure-zero condition that is not learnable by gradient descent in practice. The network converges instead to a robust piecewise-linear approximation, where local dynamics within each region are either near-marginal (rotation-like, carrying the heading forward) or, in a few regions, weakly contracting toward an embedded fixed point (3 stable nodes, 2 saddles), producing slow drift toward these discrete resting headings. This tiling is fine enough to preserve near-perfect heading decoding ($R^2\approx1$ from the full state, $R^2=0.97$ from the bitcode alone) and the resulting drift remains small within typical trial lengths ($\sim$40 timesteps).  \textbf{B}: Task accuracy as a function of the angular error threshold, for different numbers of nonlinear units, averaged across 5 seeds. As in our other threshold-ablation analyses (Table ~\ref{tab:battery_accuracy}), the choice of accuracy threshold affects how well different architectures are distinguished, where progressively tighter thresholds reveal a graded improvement in ring-attractor approximation quality with increasing nonlinearity.}
    \label{fig:head_discriminator_task}
\end{figure}

\begin{figure}[htb!]
    \centering
    \includegraphics[width=.99\textwidth]{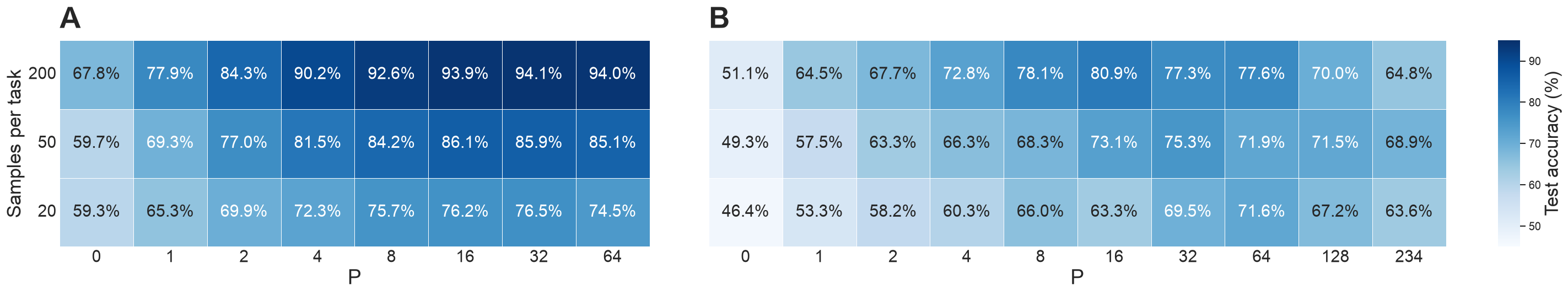}
    \caption{Test accuracy (\%) as a function of nonlinearity $P$ and training sample size per task, for \textbf{A} individual models (hidden size $M=64$) trained separately on each of 100 tasks and \textbf{B} a single flat model (hidden size $M=234$) jointly trained on all 100 tasks. Values are means over 5 seeds.}
    \label{fig:samples_vs_nonlinearity}
\end{figure}

\begin{figure}[!htb]
    \centering
    \includegraphics[width=0.99\linewidth]{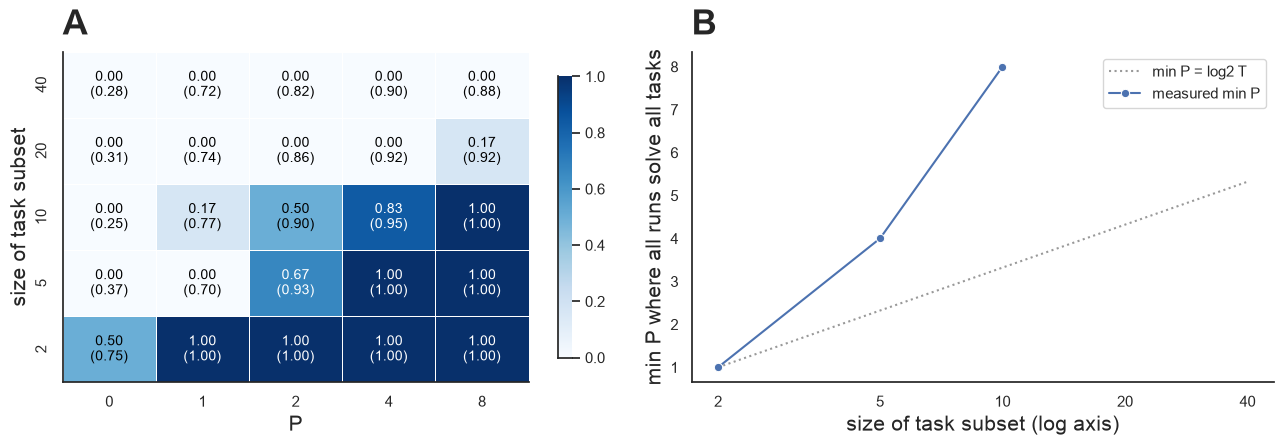}
    \caption{Individually linear tasks require nonlinear capacity when learned jointly. We constructed subsets of $T$ tasks for which every task was solved by an individually trained linear AL-RNN, and trained joint models with varying numbers of nonlinear units. \textbf{A:} Each cell reports the fraction of six runs that solved all $T$ tasks, values in parentheses show the mean fraction of tasks solved. \textbf{B}: Minimum number of nonlinear units required to solve every task in a subset reliably. The dotted line shows the idealized lower bound $\log_2 T$ obtained by assigning each task a distinct nonlinear region. The larger number required in practice likely reflects the additional constraint of shared recurrent parameterization in the AL-RNN.}
    \label{fig:nonlinear_joint}
\end{figure}

\begin{figure}[!htb]
    \centering
\includegraphics[width=0.95\linewidth]{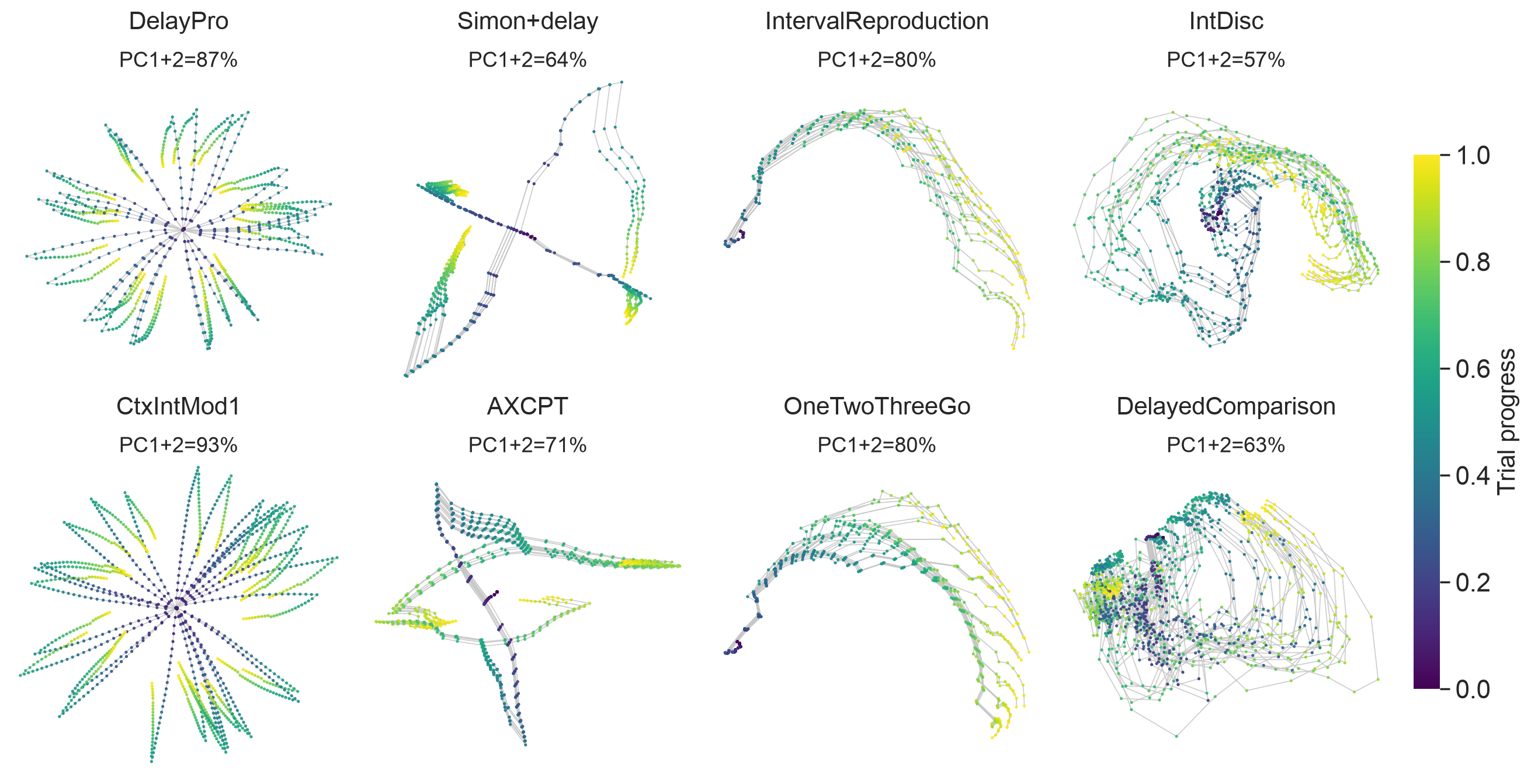}
    \caption{Four example tasks, each shown with one of its closest neighbours by geometric overlap. Tasks placed close together also tend to produce trajectories of similar shape.}
    \label{fig:similar_trajectories}
\end{figure}

\input{tables/table_robustness_features}

\input{tables/table_cross_robustness}

\begin{figure}[htb!]
    \centering
    \includegraphics[width=.99\textwidth]{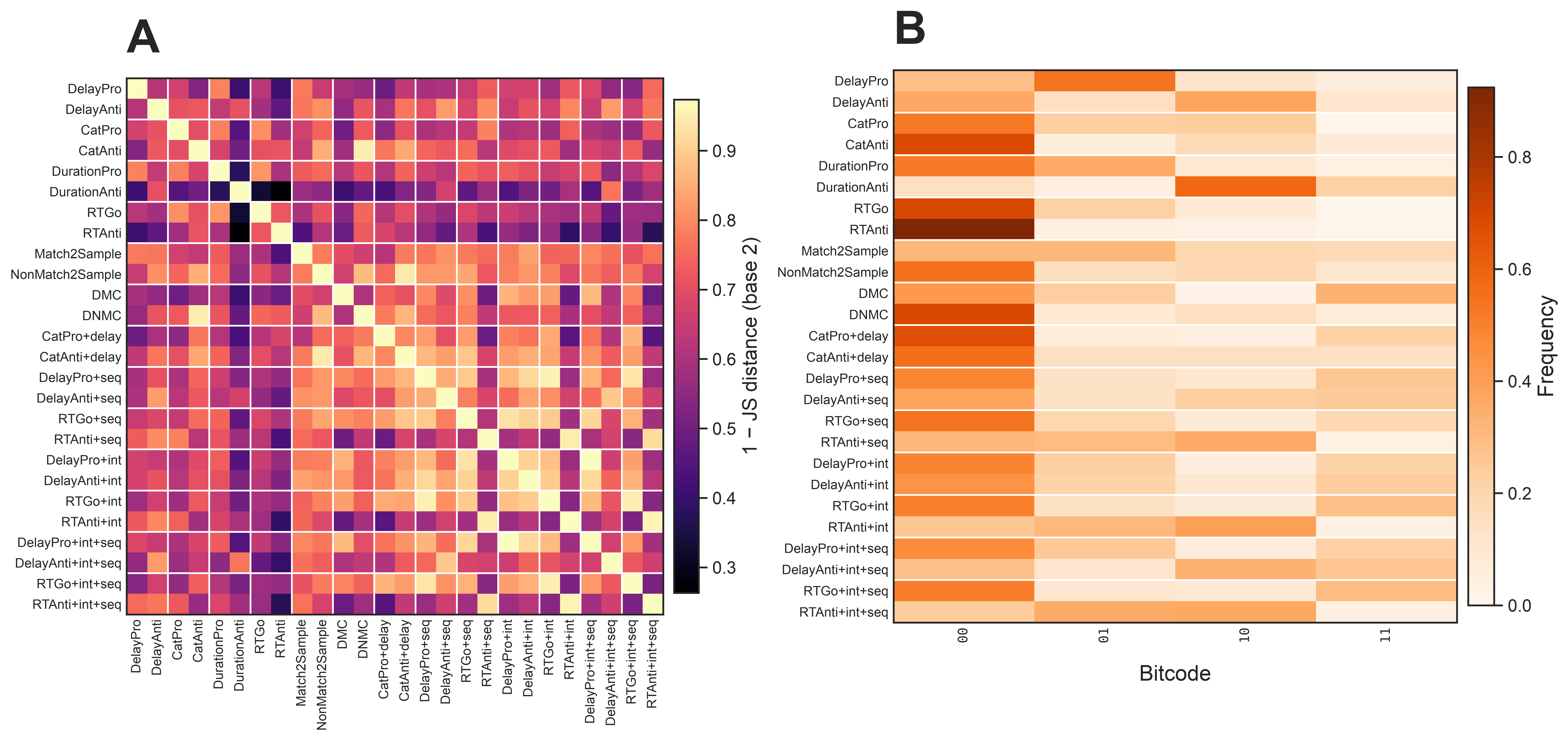}
    \caption{\textbf{A}: Linear region usage for a hierarchical AL-RNN with $P=2$, leading to $2^2=4$ linear subregions. \textbf{B}: Pro-Anti task pairs are separated in nonlinear space. Each row shows the frequency with which a task visits each linear region of the AL-RNN. Pro and anti variants share some regions but also exhibit distinct usage patterns, indicating that the model reuses shared nonlinear structure while separating computations with opposing response rules.}
    \label{fig:pro_anti_bitcodes}
\end{figure}

\pagebreak
\newpage

\input{tasks}

\end{document}

%% file: math_commands.tex
\usepackage{amsmath,amsfonts,bm}

\def\eqref#1{equation~\ref{#1}}
\def\1{\bm{1}}

\DeclareMathAlphabet{\mathsfit}{\encodingdefault}{\sfdefault}{m}{sl}
\SetMathAlphabet{\mathsfit}{bold}{\encodingdefault}{\sfdefault}{bx}{n}

%% file: tables/table_main_slim.tex
\begin{table}[!htb]
\centering
\caption{Performance of the best nonlinear model for each configuration and the corresponding linear models, with matched parameter counts and $200$ samples per task. A task counts as solved if its test accuracy reaches at least 80\% in at least 3 of the 5 seeds. $^\dagger$52 tasks are solved by at least one of the 5 seeds of the linear individual model; we refer to these as \emph{linear} tasks.
}
\label{tab:main_performance}
\begin{tabular}{lcccccc}
\toprule
Model & $M$ & $P$ & $N_\text{feat}$ & Test accuracy (\%) & Tasks solved & Parameters \\
\midrule
\multicolumn{7}{l}{\textit{Nonlinear models}} \\
LSTM & 32 & -- & 16 & 96.3 $\pm$ 0.1 & 94 & 87K \\
GRU & 32 & -- & 16 & 95.6 $\pm$ 0.7& 94 & 66K \\
Vanilla RNN & 64 & -- & 16 & 93.8 $\pm$ 0.4& 90 & 81K \\
Hierarchical AL-RNN & 64 & 32 & 16 & 96.5 $\pm$ 0.3 & 94 & 82K \\
Flat joint & 234 & 16 & -- & 80.9 $\pm$ 4.3 & 65 & 81K \\
Individual & 23 & 16 & -- & 91.3 $\pm$ 0.6 & 87 & 84K \\
\midrule
\multicolumn{7}{l}{\textit{Linear models}} \\
Hierarchical AL-RNN & 64 & 0 & 16 & 70.4 $\pm$ 0.3 & 45 & 81K \\
Flat joint & 234 & 0 & -- & 51.1 $\pm$ 0.8 & 14 & 81K \\
Individual & 23 & 0 & -- & 64.1 $\pm$ 1.2 & 42$^\dagger$ & 83K \\
\bottomrule
\end{tabular}
\end{table}

%% file: tables/table_accuracy.tex
\begin{table}[!htb]
\begin{tabular}{ll}
\toprule
Response type & Correct if \\
\midrule
Angular & $|\theta - \theta^{*}| < 36^\circ$ \\
Binary & output and target on same side of $0.5$ \\
Scalar / vector & readout error $< 0.35\,\sigma$ \\
$K$-way categorical & argmax over readout channels matches target \\
Sequence production & $|\theta_t - \theta_t^{*}| < 36^\circ$ at each step \\
Flip-flop & correct sign on every bit channel at each step \\
\bottomrule
\end{tabular}
\centering
\caption{Trial-level accuracy criteria by response type. $\theta$ denotes the angle decoded from the $(\cos\theta,\sin\theta)$ output pair; $\sigma$ the standard deviation of the task's target distribution. Single-response tasks are evaluated at the last scored timestep; sequential tasks are
scored as the fraction of scored timesteps meeting the criterion.}
\label{accuracy}
\end{table}

%% file: tables/table_accuracies.tex
\begin{table}[!htb]
\centering
\caption{Robustness of mean battery accuracy to the angular accuracy threshold, averaged across 5 seeds. Values show mean test accuracy (\%) for models with $N_{\mathrm{feat}}=16$. The $36^\circ$ threshold is used in the main text. Although absolute accuracy increases with the tolerance, the relative ordering of nonlinear capacities is largely preserved. The final column reports the change between the most and least permissive thresholds.}
\label{tab:battery_accuracy}
\begin{tabular}{lccccccccc}
\toprule
& $15^\circ$ & $20^\circ$ & $25^\circ$ & $30^\circ$ & $36^\circ$ & $40^\circ$ & $45^\circ$ & $50^\circ$ & $\Delta 50-15$ \\
\midrule
P & & & & & & & & &  \\
\midrule
1 & 72.9 & 76.5 & 79.4 & 81.6 & 83.7 & 84.9 & 86.1 & 87.1 & 14.2 \\
2 & 78.0 & 82.2 & 85.1 & 87.3 & 89.2 & 90.1 & 91.1 & 91.8 & 13.8 \\
4 & 81.6 & 85.5 & 88.2 & 90.1 & 91.7 & 92.5 & 93.3 & 93.8 & 12.2 \\
8 & 85.0 & 89.0 & 91.5 & 93.1 & 94.4 & 95.0 & 95.5 & 95.9 & 10.9 \\
16 & 89.0 & 92.4 & 94.1 & 95.2 & 95.8 & 96.2 & 96.4 & 96.6 & 7.6 \\
32 & 90.2 & 93.3 & 94.9 & 95.8 & 96.4 & 96.7 & 96.9 & 97.0 & 6.8 \\
64 & 89.0 & 92.2 & 94.1 & 95.2 & 95.9 & 96.2 & 96.4 & 96.6 & 7.6 \\
\bottomrule
\end{tabular}
\end{table}

%% file: tables/table_robustness_features.tex
\begin{table*}[!htb]
\centering
\caption{Robustness and overlap of task-similarity spaces across random seeds. Entries report mean seed-pair Pearson correlations, with $\Delta r=(r_{\max}-r_{\min})/2$ denoting half the observed range. The upper panel measures reproducibility within each similarity space, while the lower panel measures correlations between spaces. The robustness of a task representation is partly coupled to task performance, since poor performance constrains the feature vectors and dynamical activity less. On the other hand, particularly for the nonlinearity usage, we found that increasing capacity enforces less sharing between model components, and tasks are separated into increasingly distinct regions, reducing off-diagonal similarity and making the nonlinear usage matrix approach an identity-like structure. Accordingly, the cross-space agreement was generally highest at intermediate nonlinear capacity, consistent with sparse nonlinearity providing task differentiation without over-separating tasks into distinct linear regions.}
\label{tab:task_space_correlations}
\scalebox{0.85}{%
\begin{tabular}{lcccccc}
& \multicolumn{3}{c}{\textbf{$N_{\mathrm{feat}}=8$}}
& \multicolumn{3}{c}{\textbf{$N_{\mathrm{feat}}=16$}} \\
Similarity space / pair
& $P=8$ & $P=16$ & $P=32$
& $P=8$ & $P=16$ & $P=32$ \\
\midrule
\multicolumn{7}{l}{\emph{Within-space reproducibility}} \\
Hierarchical features
& $0.56\pm0.06$ & $0.63\pm0.06$ & $0.64\pm0.06$
& $0.76\pm0.02$ & $0.78\pm0.04$ & $0.82\pm0.01$ \\
Task geometry
& $0.67\pm0.04$ & $0.74\pm0.04$ & $0.72\pm0.04$
& $0.77\pm0.03$ & $0.78\pm0.02$ & $0.82\pm0.02$ \\
Nonlinear usage
& $0.66\pm0.06$ & $0.78\pm0.05$ & $0.85\pm0.05$
& $0.71\pm0.07$ & $0.76\pm0.03$ & $0.81\pm0.05$ \\
\midrule
\multicolumn{7}{l}{\emph{Between-space similarity}} \\
Geometry--features
& $0.69\pm0.04$ & $0.69\pm0.02$ & $0.65\pm0.05$
& $0.65\pm0.04$ & $0.65\pm0.04$ & $0.63\pm0.03$ \\
Geometry--nonlinearity
& $0.54\pm0.04$ & $0.70\pm0.04$ & $0.53\pm0.03$
& $0.56\pm0.07$ & $0.68\pm0.05$ & $0.52\pm0.03$ \\
Features--nonlinearity
& $0.37\pm0.08$ & $0.47\pm0.04$ & $0.32\pm0.04$
& $0.37\pm0.11$ & $0.47\pm0.07$ & $0.34\pm0.05$ \\
\bottomrule
\end{tabular}%
}
\end{table*}

%% file: tables/table_cross_robustness.tex
\begin{table}[t]
\centering
\caption{Seed reproducibility and cross-architecture agreement for task representations across 5 seeds. The nonlinear feature space is only feasible for the AL-RNN, and hence omitted here. Models with settings as in Table \ref{tab:main_performance} ($M=64$, $N_{feat}=16$).}
\label{tab:cross_architecture_task_spaces}
\begin{tabular}{lcc}
\toprule
Model & Geometry & Task features \\
\midrule
LSTM seed consistency & 0.81 & 0.76 \\
GRU seed consistency  & 0.78 & 0.74 \\
\midrule
AL-RNN--LSTM feature similarity & \multicolumn{2}{c}{0.53} \\
AL-RNN--GRU feature similarity  & \multicolumn{2}{c}{0.57} \\
\bottomrule
\end{tabular}
\end{table}

%% file: tasks.tex
\newcommand{\taskentry}[3]{#1 & #3 & #2 \\\addlinespace}

\newenvironment{modtable}{%
  \par\smallskip\noindent\small\setlength{\tabcolsep}{4pt}%
  \begin{tabular}{@{}>{\raggedright\arraybackslash}p{0.21\linewidth}
                     >{\raggedright\arraybackslash}p{0.61\linewidth}
                     >{\raggedright\arraybackslash}p{0.13\linewidth}@{}}
  \toprule
  \textbf{Modifier} & \textbf{Description} & \textbf{Readout} \\
  \midrule}
 {\bottomrule\end{tabular}\par\smallskip}

\clearpage
\section{Task battery}
\label{tasks_list}
%All $K = 100$ tasks share the interface of Sec.~\ref{sec:interface}. 

The task descriptions and names were adapted from established paradigms in cognitive psychology and systems neuroscience where available. Citations indicate the source paradigm and conceptual inspiration. Our implementations are standardized for the shared low-dimensional interface and therefore do not always reproduce the original experimental stimuli or modalities exactly. For example, auditory, visual, and motor variables are represented by abstract low-dimensional input and output channels rather than their natural sensory or behavioural formats.

Observations are zero-padded to a common input dimension, targets live in a fixed $5$-dimensional readout, and a masked mean-squared error restricts the loss to the behaviourally relevant timesteps of each trial. Trials are organised into phases (\emph{context}, \emph{stimulus}, \emph{delay}, \emph{response}) whose durations are resampled per trial, and a fixation signal is maintained until the response window.

Table~\ref{tab:tasks} lists the $66$ distinct paradigms and Sec.~\ref{modifiers} the four modifiers that expand them to the $100$ task identities used in the experiments, based on the Mod-Cog suite \citep{khona2023winninglotteryneuralconnectivity}. Throughout, \emph{angular} denotes a unit vector $(\cos\theta,\sin\theta)$ scored by angular error, \emph{binary} a thresholded decision, \emph{scalar} a continuous magnitude, \emph{vector} two continuous coordinates, \emph{categorical} an $\arg\max$ over $K$ alternatives, and \emph{bits} the signs of $n$ outputs. A star ($^\star$) marks readouts scored at every response timestep rather than only at the report step.

\subsection{Modifiers}
\label{modifiers}

The paradigms of Table~\ref{tab:tasks} are expanded by four modifiers, each of which wraps a base task into a separate task identity with its own embedding $\bm{p}^{(k)}$, giving $66 + 34 = 100$ tasks in total. The bases they apply to are $\{$GoNogo, PerceptualDM, CatPro, CatAnti, NoiseCleaner, MultiSens, Stroop, Flanker, Simon, TaskSwitch$\}$ for $+$delay, and the eight single-target angular tasks $\{$DelayPro, DelayAnti, NoiseCleaner, MultiSens, PerceptualDM, RTGo, RTAnti, SpatialWMRetro$\}$ for $+$seq, $+$int and their composition.

\begin{modtable}
\taskentry{$+$delay}{as base}{A blank retention gap (fixation on, stimulus off, unscored) is spliced in immediately before the response window, forcing the pre-response state to be held across the delay.}
\taskentry{$+$seq}{angular$^\star$}{The reported direction drifts at a cued angular speed $\omega$ over the response window, so a static angular response becomes a rotating output trajectory.}
\taskentry{$+$int}{angular}{A delay of variable length is inserted before the response during which the remembered direction rotates at $\omega$, so the reported angle is $\theta + \omega T_{\text{delay}}$ and angular velocity must genuinely be integrated.}
\taskentry{$+$int$+$seq}{angular$^\star$}{Composition of the two: the memory rotates across the inserted delay \emph{and} the output moves during the response window.}
\end{modtable}

\small
\setlength{\tabcolsep}{4pt}
\begin{longtable}{@{}>{\raggedright\arraybackslash}p{0.21\linewidth}
                     >{\raggedright\arraybackslash}p{0.61\linewidth}
                     >{\raggedright\arraybackslash}p{0.13\linewidth}@{}}
\caption{The $66$ paradigms of the task battery}\label{tab:tasks}\\
\toprule
\textbf{Task} & \textbf{Description} & \textbf{Readout} \\
\midrule
\endfirsthead
\multicolumn{3}{@{}l}{\emph{Table~\ref{tab:tasks}, continued}}\\
\toprule
\textbf{Task} & \textbf{Description} & \textbf{Readout} \\
\midrule
\endhead
\midrule
\multicolumn{3}{r@{}}{\emph{continued on next page}}\\
\endfoot
\bottomrule
\endlastfoot

\taskentry{DelayPro}{angular}{A direction cue is presented, followed by a delay phase. After delay, the network reports it once the fixation cue is removed \citep{yang_task_2019}.}
\taskentry{DelayAnti}{angular}{As DelayPro, but the reported direction is opposite to the cued one \citep{yang_task_2019,MunozEverling2004}.}
\taskentry{RTGo}{angular}{The stimulus direction is reported as soon as it appears, with no intervening delay \citep{yang_task_2019}.}
\taskentry{RTAnti}{angular}{As RTGo, but reporting the direction opposite the stimulus \citep{yang_task_2019,MunozEverling2004}.}
\taskentry{PerceptualDM}{angular}{Noisy directional evidence is presented at varying coherence and the net direction must be integrated out of the noise within a given tolerance level \citep{Britten1992}.}
\taskentry{NoiseCleaner}{angular}{A true direction must be estimated from a stream of noisy angular samples within a given tolerance.}
\taskentry{MultiSens}{angular}{Two modalities are shown simultaneously with independent strengths. After a delay period, the response is the direction of their vector sum \citep{yang_task_2019}.}
\taskentry{CtxIntMod1}{angular}{Integrate modality 1 and ignore modality 2, as instructed by a context cue shown at the start \citep{mante_context-dependent_2013, yang_task_2019}.}
\taskentry{CtxIntMod2}{angular}{Integrate modality~2 and ignore modality~1 \citep{mante_context-dependent_2013, yang_task_2019}.}
\taskentry{PriorIntegration}{angular}{A prior cue indicating the likely direction is followed by noisy evidence of varying strength. The network reports the true direction.}
\taskentry{SpatialWMRetro}{angular}{Two locations are presented in a sequence, followed by a delay and retro-cue. The network reports the cued location \citep{Funahashi1989,GriffinNobre2003}.}
\taskentry{WMUpdating}{angular}{A stream of six items, each tagged with one of two categories, is followed by a query naming a category. The network reports the most recent item of that category \citep{Yntema1963,Miyake2000}.}
\taskentry{PosnerCueing}{angular}{A spatial cue indicates the likely target direction. After a delay, a noisy target appears and the network reports its true direction \citep{Posner1980}.}
\taskentry{CopyTask}{angular$^\star$}{A sequence of symbols is presented one per step and must be immediately  reproduced in order after input ends.}
\taskentry{HeadDirection}{angular$^\star$}{An initial heading is cued and a stream of angular velocities must be integrated into a running heading reported at every step \citep{Kim2017,CuevaWei2018}.}
\taskentry{SineGen}{angular$^\star$}{A cued frequency must be turned into a sustained, amplitude-stable oscillation, i.e.\ a tunable limit cycle \citep{sussillo_opening_2013}.}
\taskentry{AntiReach1D}{scalar}{A target appears at scalar position $x$ on a line and the network must reach to $-x$ \citep{MunozEverling2004}.}
\taskentry{SoundLocalization}{scalar}{Source azimuth must be recovered from a noisy level difference across two "ear" inputs \citep{KnudsenKonishi1978}.}
\taskentry{IntervalReproduction}{scalar}{Two flashes mark an interval; the network reports its length as a scalar once the second flash ends \citep{JazayeriShadlen2010}.}
\taskentry{OneTwoThreeGo}{scalar}{Three equally spaced flashes mark a repeated interval. The network reports its length as a scalar after the third flash \citep{Egger2019}.}
\taskentry{PathIntegration1D}{scalar}{A stream of 1D velocities from a fixed origin; report the final position without further input \citep{CuevaWei2018,Seung1996}.}
\taskentry{Numerosity}{scalar}{A number of discrete pulses is presented and, after a go cue, the count is reported as a continuous quantity \citep{NiederDiesterTudusciuc2006}.}
\taskentry{PathIntegration2D}{vector}{As PathIntegration1D, with 2D velocities (vx, vy). Report the final (x, y) position \citep{CuevaWei2018,Banino2018}.}
\taskentry{GoNogo}{binary}{Respond only if the stimulus amplitude exceeds a threshold, otherwise maintain fixation and withhold the response.}
\taskentry{StopSignal}{binary}{A go stimulus appears, followed on a fraction of trials by a stop signal. Respond unless a stop signal occurred \citep{LoganCowan1984,Hanes1998}.}
\taskentry{CatPro}{binary}{Classify the stimulus direction according to which half of the ring it falls in \citep{FreedmanAssad2006}.}
\taskentry{CatAnti}{binary}{As CatPro with the category-to-response mapping reversed \citep{FreedmanAssad2006}.}
\taskentry{PoissonClicks}{binary}{Two Poisson click streams arrive on separate input lines. The network has to report which side delivered more clicks \citep{Brunton2013,Hanks2015}.}
\taskentry{SignalDetection}{binary}{A weak constant signal is present in a noisy stream on half of the trials. The network reports whether the signal was present or not \citep{GreenSwets1966}.}
\taskentry{BackwardMasking}{binary}{A brief category signal is immediately followed by a high-noise mask on the same input line. The network reports the category \citep{BreitmeyerGanz1976}.}
\taskentry{Match2Sample}{binary}{The network reports whether a test direction matches the sample held across a delay \citep{yang_task_2019}.}
\taskentry{NonMatch2Sample}{binary}{As Match2Sample, responding positively to non-match instead \citep{yang_task_2019}.}
\taskentry{DMC}{binary}{A sample direction is followed, after a delay, by a second direction. The network reports whether both fall in the same half of the ring \citep{FreedmanAssad2006,yang_task_2019}.}
\taskentry{DNMC}{binary}{As DMC, but responding positively for different halves \citep{FreedmanAssad2006,yang_task_2019}.}
\taskentry{DMSDistractor}{binary}{Delayed match-to-sample in which an irrelevant distractor direction appears during the delay and must be ignored \citep{MillerEtAl1996}.}
\taskentry{Sternberg}{binary}{A set of items is presented, followed by a delay and single probe. The network reports whether the probe was in the set \citep{Sternberg1966}.}
\taskentry{PairedAssoc}{binary}{A cue direction is followed, after a delay, by a second cue. The network reports whether the second is the designated associate of the first \citep{SakaiMiyashita1991}.}
\taskentry{ChangeDetect}{binary}{A sample array is shown and, after a delay, followed by a test array. The network  reports whether any item has changed \citep{LuckVogel1997}.}
\taskentry{SerialOrderMemory}{binary}{Report whether one probe item occurred before another in the previously presented list \citep{Hacker1980}.}
\taskentry{DelayedComparison}{binary}{Two noisy scalar magnitudes separated by a delay. Report whether the first is stronger \citep{Romo1999}.}
\taskentry{DurationPro}{binary}{Judge whether the interval between two flashes is short or long relative to a learned midpoint \citep{ChurchDeluty1977}.}
\taskentry{DurationAnti}{binary}{As DurationPro with the response mapping reversed \citep{ChurchDeluty1977}.}
\taskentry{IntDisc}{binary}{Two stimuli are presented, separated by a delay. The network reports whether the first lasted longer than the second.}
\taskentry{TemporalBisection}{binary}{A single stimulus duration is classified as short or long about a fixed, learned boundary \citep{ChurchDeluty1977,LeonShadlen2003}.}
\taskentry{TemporalOrderJudgment}{binary}{Two pulses separated by a variable onset asynchrony arrive on different input lines. The network reports which came first \citep{HirshSherrick1961}.}
\taskentry{SimultaneityJudgment}{binary}{Report whether two pulses were simultaneous or separated by a small onset asynchrony \citep{HirshSherrick1961}.}
\taskentry{GapDetection}{binary}{An ongoing stimulus is presented and the network reports whether a brief gap interrupted it \citep{Plomp1964}.} 
\taskentry{Stroop}{binary}{Two features are shown at once and a rule cue selects which to report. On incongruent trials the features disagree \citep{Stroop1935}.}
\taskentry{Flanker}{binary}{A target value and an irrelevant flanker value are shown simultaneously on separate input lines. The network reports the target
\citep{EriksenEriksen1974}.}
\taskentry{Simon}{binary}{Respond according to stimulus identity while ignoring an irrelevant location that may map to the opposite response \citep{Simon1969}.}
\taskentry{TaskSwitch}{binary}{A rule cue selects whether to judge the sign or the magnitude of the same scalar stimulus \citep{Sohn2000,Sakai2008}.}
\taskentry{AXCPT}{binary}{A cue (A or B) is followed, after a delay, by a probe (X or Y). The network responds only to the A-then-X sequence \citep{ServanSchreiber1996,Cohen1999}.}
\taskentry{DelayedCueReversal}{binary*}{A rule cue may appear early and is withdrawn before the stimulus. The network reports the stimulus sign, inverted if the cue appeared.}
\taskentry{VisualSearch}{binary}{A target feature is cued and the network reports whether it is present in a sequentially presented array \citep{TreismanGelade1980}.}
\taskentry{AttentionalBlink}{binary}{A rapid stream contains two flagged targets at variable lag. The network reports the category of the second \citep{RaymondShapiroArnell1992}.}
\taskentry{OddballDetection}{binary}{A stream of repeated cues is presented and the network reports whether a single cue broke the regularity \citep{Naatanen1978,Ulanovsky2003}.}
\taskentry{HierReason}{binary}{A scalar stimulus is presented with two cues: one selects whether to judge its sign or its magnitude, the other whether to invert that judgment. The network reports the result \citep{BadreDEsposito2007,BadreDEsposito2009}.}
\taskentry{ConditionalReasoning}{binary}{An antecedent and a consequent are presented and the network reports whether the case is consistent with the material conditional, i.e.\ detects the falsifying combination \citep{Wason1968,JohnsonLairdTagart1969}.}
\taskentry{RelationalMatch}{binary}{Two pairs of items are presented in a sequence, each either same or different, and the network reports whether both pairs share the same relation \citep{Fagot2001}.}
\taskentry{MentalRotation}{binary}{A sample direction is followed by a delay and a probe, with the rotation amount cued from trial onset. The network reports whether the probe equals the sample rotated by the cued amount \citep{ShepardMetzler1971,Georgopoulos1989}.}
\taskentry{SymbolCategory}{binary}{Two directional symbols are presented in sequence; the network reports whether the first belongs to a fixed category, with the second acting as a distractor.}
\taskentry{DiscriminationOptOut}{categorical}{Noisy scalar evidence of variable strength is presented. The network chooses left or right when the underlying signal is strong and a third "opt-out" option when it is weak.}
\taskentry{PluralityVote}{categorical}{A stream of six symbols, each one of three categories, is presented and followed by a delay. The network reports the most frequent category.}
\taskentry{OddOneOut}{categorical}{$K$ scalar items are shown at once and exactly one deviates from the common level (up or down). The network reports which one.}
\taskentry{FlipFlop3}{3 bits$^\star$}{Three bits, each set by a brief $\pm1$ pulse, must be held and reported at every timestep \citep{sussillo_opening_2013}.}
\taskentry{FlipFlop2}{2 bits$^\star$}{As FlipFlop3 with two bits, giving four stable configurations.}

\end{longtable}
\normalsize
\setlength{\tabcolsep}{6pt}

%\section{References for task paradigms}
%\label{task_references}

%A citation in Table~\ref{tab:tasks} is given only where both the paradigm and its source were verified against the task implementation and the published record (all DOIs resolved through Crossref). A \emph{(???)} marks a task for which no verified original-paradigm source has been established yet.

%\begin{thebibliography}{99}

%\end{thebibliography}

%\subsection*{Unresolved citation placeholders}

%The following tasks carry a \emph{(???)} in Table~\ref{tab:tasks} because no original-paradigm source has been verified against their implementation: AntiReach1D, GoNogo, CatPro, CatAnti, PerceptualDM, NoiseCleaner, MultiSens, CtxIntMod1, CtxIntMod2, PriorIntegration, MetacognitiveOptOut, DelayedComparison, CopyTask, DurationPro, DurationAnti, IntDisc, TemporalOrderJudgment, SimultaneityJudgment, GapDetection, ProspectiveMemory, RelationalMatch, PluralityVote, and OddOneOut.